\documentclass[10pt,journal,compsoc]{IEEEtran}
\usepackage{amsmath}
\usepackage{amsfonts}
\usepackage{bbding}
\usepackage{amsmath}
\usepackage{amsfonts}
\usepackage[utf8]{inputenc} 
\usepackage[T1]{fontenc}    
\usepackage{hyperref}       
\usepackage{url}            
\usepackage{booktabs}       
\usepackage{amsfonts}       
\usepackage{nicefrac}       
\usepackage{microtype}      
\usepackage{amsthm,amsmath,amssymb}
\usepackage{mathrsfs}
\usepackage{graphicx}
\usepackage{subfigure}
\usepackage{color}
\usepackage{wrapfig}
\usepackage{bbding}
\usepackage{float}
\usepackage{multirow}

\usepackage{graphicx}

\usepackage{graphicx}
\usepackage{amsmath}
\usepackage{amssymb}
\usepackage{amscd}
\usepackage{booktabs}
\usepackage{algorithm}
\usepackage{algpseudocode}
\usepackage{amsthm}
\usepackage{textcomp}
\usepackage{amsthm,amsmath,amssymb}
\usepackage{mathrsfs}
\usepackage{graphicx}
\usepackage{wrapfig}
\usepackage{bbding}
\usepackage{float}
\usepackage{mathrsfs}
\usepackage{bbm}
\usepackage{ragged2e}

\usepackage[accsupp]{axessibility}

\ifCLASSOPTIONcompsoc
  \usepackage[nocompress]{cite}
\else
  \usepackage{cite}
\fi
\ifCLASSINFOpdf
\else
\fi

\begin{document}
%
\title{On-the-fly Modulation for \\ Balanced Multimodal Learning}
%
%
%
%

\author{
        Yake Wei,
        Di Hu,
        Henghui Du,
        Ji-Rong Wen,~\IEEEmembership{Senior~Member,~IEEE}
\IEEEcompsocitemizethanks{

\IEEEcompsocthanksitem Y. Wei, D. Hu (corresponding author), H. Du, and J-R. Wen are with the Gaoling School of Artificial Intelligence, and Beijing Key Laboratory of Big Data Management and Analysis Methods, Renmin University of China, Beijing 100872, China.\protect\\
E-mail: \{yakewei, dihu, cserdu, jrwen\}@ruc.edu.cn
}
}

\IEEEtitleabstractindextext{%
\begin{abstract}
Multimodal learning is expected to boost model performance by integrating information from different modalities. However, its potential is not fully exploited because the widely-used joint training strategy, which has a uniform objective for all modalities, leads to imbalanced and under-optimized uni-modal representations. Specifically, we point out that there often exists modality with more discriminative information, \emph{e.g.,} vision of \textit{playing football} and sound of \textit{blowing wind}. They could dominate the joint training process, resulting in other modalities being significantly under-optimized. To alleviate this problem, we first analyze the under-optimized phenomenon from both the feed-forward and the back-propagation stages during optimization. Then, On-the-fly Prediction Modulation (OPM) and On-the-fly Gradient Modulation (OGM) strategies are proposed to modulate the optimization of each modality, by monitoring the discriminative discrepancy between modalities during training. Concretely, OPM weakens the influence of the dominant modality by dropping its feature with dynamical probability in the feed-forward stage, while OGM mitigates its gradient in the back-propagation stage. In experiments, our methods demonstrate considerable improvement across a variety of multimodal tasks. These simple yet effective strategies not only enhance performance in vanilla and task-oriented multimodal models, but also in more complex multimodal tasks, showcasing their effectiveness and flexibility. The source code is available at \url{https://github.com/GeWu-Lab/BML_TPAMI2024}.
\end{abstract}

}

\maketitle

\maketitle

\IEEEdisplaynontitleabstractindextext

%
\IEEEpeerreviewmaketitle

\IEEEraisesectionheading{\section{Introduction}\label{sec:introduction}}

\IEEEPARstart{P}{e}ople perceive the surrounding world by comprehensively integrating multiple senses, including vision, hearing, and touch. This process is known in cognitive neuroscience as multi-sensory integration~\cite{gazzaniga2006cognitive}. Inspired by this phenomenon, multimodal data, collected from multiple sensors, has raised attention in the machine learning field, and accordingly multimodal learning has witnessed significant advances in these years. The research community has improved the performance of traditional uni-modal tasks by incorporating additional modalities and has also begun tackling new, challenging problems~\cite{wei2022learning}, such as multimodal action recognition~\cite{kazakos2019epic,gao2020listen}, multimodal semantic segmentation~\cite{choudhury2018segmentation,cao2021shapeconv} and audio-visual event localization~\cite{tian2018audio}.

Multimodal models are expected to surpass their uni-modal counterparts since they take data containing information from multiple views. In most cases, it does achieve this intention, but sometimes goes the contrary: the multimodal model can be inferior to the uni-modal one~\cite{wang2020makes}. In recent studies, some researchers claimed that different modalities could perform dis-similarly in the optimization process. For instance, the audio modality tends to converge with a faster learning pace in the video recognition task, compared with the visual one~\cite{wang2020makes}. This discrepancy makes it challenging for multimodal models to effectively learn from all modalities simultaneously under a uniform joint training objective~\cite{sun2021learning,huang2022modality}. As a result, the potential of multimodal models can be limited by the difference in the learning status of different modalities, and then fail to outperform uni-modal counterparts.
To cope with this issue, some studies depending on added uni-modal classifiers or additional training for the specific modality are proposed~\cite{wang2020makes,wu2022characterizing}, but they inevitably introduce extra training efforts.

\begin{figure*}[h]
 \centering
    \subfigure []{
    \includegraphics[width=0.31\linewidth]{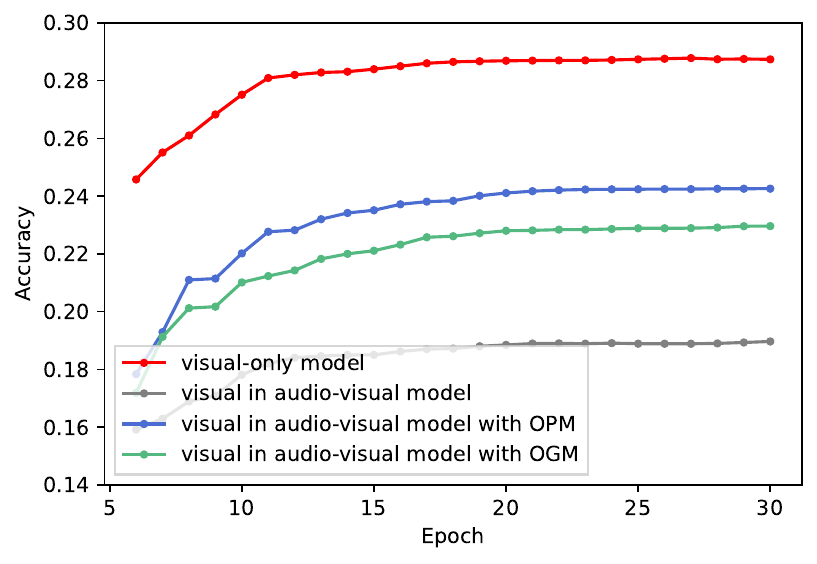}
    \label{fig-teaser-visual}
    }
    \subfigure[]{
   \includegraphics[width=0.31\linewidth]{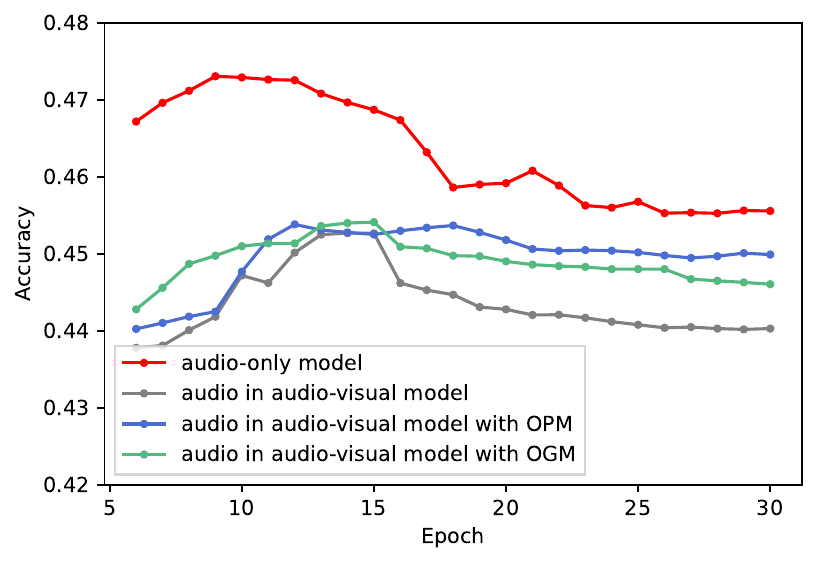}
   
    \label{fig-teaser-audio}
    }
    \subfigure[]{
    \includegraphics[width=0.31\linewidth]{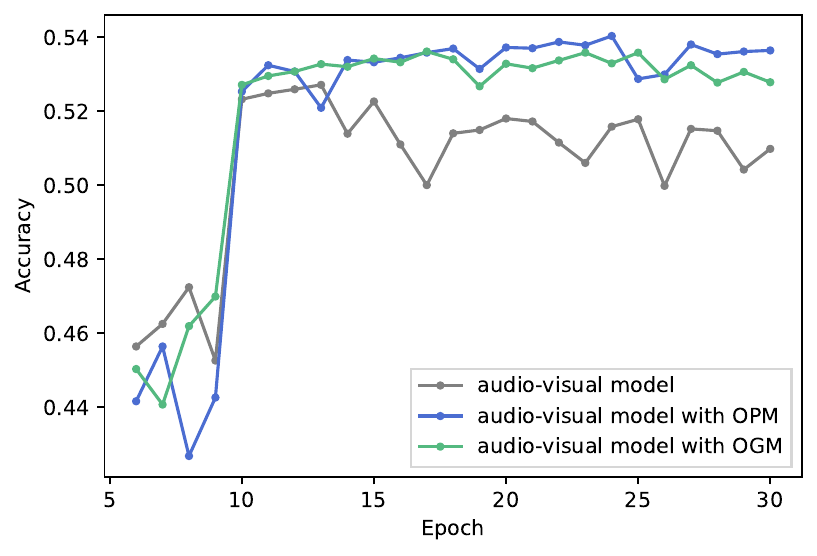}
    \label{fig-teaser-av}
    }
    \vspace{-1em}
    \caption{\textbf{Performance of individually trained uni-modal model, jointly trained multimodal model and jointly trained multimodal model with our proposed OPM and OGM strategies respectively on the VGGSound dataset.} (a) Performance of visual modality. (b) Performance of audio modality. (c) Performance of audio-visual modalities. Best viewed in color. The training of our OPM and OGM methods exactly aligns with the applied audio-visual model. To provide more representative observation, here jointly trained multimodal use concatenation fusion, which is widely-used, and simple-but-strong. In Appendix D, we also extend these experiments to more complex CentralNet~\cite{vielzeuf2018centralnet} multimodal framework.}
    \vspace{-1em}
    \label{fig-teaser} 
\end{figure*}

Beyond the failure cases of multimodal joint learning, we note that even when multimodal models outperform their uni-modal counterparts, they still fail to fully harness the potential of multiple modalities. As demonstrated in~\autoref{fig-teaser}, we conducted experiments on the VGGSound dataset~\cite{chen2020vggsound} to assess the quality of uni-modal encoders in jointly trained multimodal models. Our results show that the jointly trained multimodal models perform better than the uni-modal models, which is expected. However, when examining the performance of uni-modal encoders within these multimodal models\footnote{Here the audio-only and visual-only models are trained individually. To evaluate the quality of uni-modal encoder in jointly trained audio-visual models at a certain epoch, we first take the uni-modal encoder in audio-visual models at this epoch, then freeze its parameters and fine-tune a new uni-modal classifier. Finally, the accuracy of the uni-modal encoder in audio-visual models at a certain epoch is obtained.}, we discover that they are under-optimized compared to the corresponding solely trained uni-modal models. For example, in~\autoref{fig-teaser-visual}, during the whole training process, performance of visual-only model (red line) is better than visual encoder in audio-visual model (gray line). Moreover, \emph{the under-optimized degrees of different modalities are imbalanced, and one modality is clearly worse learnt than others}. As shown in~\autoref{fig-teaser-visual} and~\autoref{fig-teaser-audio}, the quality of visual encoder in the audio-visual model has a more clear drop, compared with the audio modality. Overall, these interesting observations demonstrate that the uni-modal representation is under-optimized with an imbalanced degree in the joint training multimodal model.

The reason could be that, for the multimodal dataset, there often exists a dominant modality~\cite{parida2020coordinated} with the better discriminative ability (\emph{e.g.,} vision of \textit{drawing} and sound of \textit{wind blowing}), which tends to be favored during training and consequently suppress the learning of others. For instance, as illustrated in~\autoref{fig-teaser}, the visual encoder in the audio-visual model is more remarkably under-optimized. This observation is consistent with the fact that the curated sound-oriented dataset, VGGSound, has a preference for the audio modality. This preference within the dataset would result in one modality often being more discriminative, causing the observed imbalanced learning problem among modalities.

To alleviate this problem, we first analyze the imbalanced learning phenomenon in the jointly trained multimodal model from both the feed-forward and back-propagation stages. In the feed-forward stage, the modality with more discriminative information often determines the model output and dominates the prediction.
Subsequently, it also lowers the joint loss more, limiting the gradient of other modalities in the back-propagation stage. These observations cause the imbalanced learnt situation between modalities. To ease this issue, we propose to control the optimization of each modality via two on-the-fly modulation methods: \emph{On-the-fly Prediction Modulation} (OPM) and \emph{On-the-fly Gradient Modulation} (OGM). These two strategies respectively target the feed-forward and back-propagation stages. Specifically, by monitoring the discriminative discrepancy between modalities during training, OPM drops the feature of dominant modality with dynamical probability while OGM on-the-fly mitigates its gradient, thereby improving the learning of the less discriminative modality. As shown in~\autoref{fig-teaser}, the previously worse learnt visual modality of the VGGSound dataset shows marked improvement after applying either OPM or OGM (blue and green lines in~\autoref{fig-teaser-visual} and~\autoref{fig-teaser-audio}). The performance of the dominant audio modality also benefits. Furthermore, our methods notably enhance the overall multimodal model with the joint training strategy (see~\autoref{fig-teaser-av}). To thoroughly demonstrate the effectiveness and versatility of OPM and OGM, we evaluate their performance across various multimodal tasks, achieving consistent improvement. Moreover, our modulation methods help improve multimodal representation by enhancing the learning of each modality.

Our previous conference paper~\cite{peng2022balanced} has exposed the imbalanced learning phenomenon in the back-propagation stage and achieved considerable performance with the proposed on-the-fly gradient modulation method. In this paper, we provide a more systematic analysis of the imbalanced learning problem from both the feed-forward and the back-propagation stages, and further introduce the on-the-fly prediction modulation method, which focuses on the feed-forward stage. These two strategies are designed to holistically consider the stages of optimization. Moreover, we additionally provide an analysis that how our improvement in representation contributes to better model performance. In addition, we further expand our methods to address more complex cross-modal interactions and a broader range of multimodal tasks. In experiments, we conduct extensive evaluation across a diverse set of modalities, encompassing various numbers and types. A wide range of fine-grained analysis and ablation studies are also conducted to validate our methods thoroughly. Overall, our main contributions are as follows:
\begin{itemize} 
\item We observe and analyze the imbalanced learning phenomenon that the uni-modal encoders in multimodal model are imbalanced under-optimized and one modality could be worse learnt than others during training, from both the feed-forward and the back-propagation stages.
\item OPM and OGM methods are proposed to ease imbalanced learning problem by controlling the optimization of each modality adaptively in the feed-forward and the back-propagation stages, respectively. 
\item OPM and OGM can be equipped with various multimodal tasks and models, demonstrating their promising effectiveness and versatility in diverse multimodal learning scenarios.
\end{itemize}

\section{Related Works}

\subsection{Multimodal learning}
Multimodal learning, which integrates information from multiple modalities, has been attracting increasing attention due to the growing amount of multimodal data. This data naturally contains correlated information from diverse sources. Recently, the field of multimodal learning has witnessed rapid development. On the one hand, multimodal modalities are used to enhance the performance of existing uni-modal tasks, such as multimodal action recognition~\cite{simonyan2014two,gao2020listen,kazakos2019epic} and audio-visual speech recognition~\cite{potamianos2004audio,hu2016temporal}. On the other hand, more researchers also begin to explore and solve new multimodal problems and challenges, like audio-visual event localization~\cite{tian2018audio,lin2019dual} and multimodal question answering~\cite{ilievski2017multimodal}. To efficiently learn and integrate multiple modalities, most multimodal methods tend to use the joint training strategy, which optimizes different modalities with a uniform learning objective. However, this approach may not fully exploit all modalities, causing some to be less effectively learned than others. This can prevent multimodal models from achieving their expected performance, even though they are superior to their uni-modal counterparts. In this paper, we propose on-the-fly modulation methods to improve the joint learning of multimodal models via dynamically controlling the uni-modal optimization.

\subsection{Imbalanced multimodal learning}
The multimodal model is expected to outperform its uni-modal counterpart since it takes data containing information from multiple views. But the widely used joint training multimodal model does not always work well based on existing studies~\cite{wang2020makes}, which prompts researchers to investigate the reasons. Recent studies point out that the jointly trained multimodal model cannot effectively improve the performance with more information as expected due to the discrepancy between modalities~\cite{wang2020makes,winterbottom2020modality,sun2021learning,huang2022modality,wu2022characterizing}. Wang et al.~\cite{wang2020makes} found that multiple modalities often converge and generalize at different rates, thus training them jointly with a uniform learning objective is sub-optimal, leading to the multimodal model sometimes is inferior to the uni-modal ones. Also, Winterbottom et al.~\cite{winterbottom2020modality} indicated an inherent bias in the TVQA dataset towards the textual subtitle modality. Besides the empirical observation, Huang et al.~\cite{huang2022modality} further theoretically proved that the jointly trained multimodal model cannot efficiently learn features of all modalities, and only a subset of them can capture sufficient representation. They called this process ``Modality Competition''.
In the recent past, several methods have emerged attempting to alleviate this problem~\cite{wang2020makes,du2021improving,wu2022characterizing,wei2024innocent}. Wang et al.~\cite{wang2020makes} proposed to add additional uni-modal loss functions besides the original multimodal objective to balance the training of each modality. Du et al.~\cite{du2021improving} utilized the well-trained uni-modal encoders to improve the multimodal model by knowledge distillation. Wu et al.~\cite{wu2022characterizing} measured the speed at which the model learns from one modality relative to the
other modalities, and then proposed to guide the model to learn from previously underutilized modalities. Wei et al.~\cite{wei2024enhancing} introduced a Shapley-based sample-level modality valuation metric, to observe and alleviate the fine-grained modality discrepancy. Wei et al.~\cite{wei2024diagnosing} further considered the possible limited capacity of modality and utilized the re-initialization strategy to control uni-modal learning. Differently, Yang et al.~\cite{yang2024Quantifying} focused on the influence of imbalanced multimodal learning on multimodal robustness, and proposed a robustness enhancement strategy. While these methods have improved multimodal learning, a comprehensive analysis of imbalanced multimodal learning is still lacking. In this paper, we begin with a systematic analysis of both the feed-forward and back-propagation stages to understand how multimodal discrepancies impact training. Based on the analysis, we propose to alleviate these issues by adaptively controlling the optimization of each modality without introducing additional modules.

\subsection{Modality dropout}
In our OPM method, we adaptively drop the feature of the dominant modality during training to enhance the learning of the remaining modalities. Following the regularization technique dropout~\cite{srivastava2014dropout}, different network dropout strategies are proposed and show their effectiveness~\cite{rennie2014annealed,li2016improved,morerio2017curriculum}. In recent years, the idea of dropout has been transferred into the multimodal learning area to drop modalities during training~\cite{neverova2015moddrop,li2016multi,hussen2020modality, de2020input,xiao2020audiovisual}. Neverova et al.~\cite{neverova2015moddrop} proposed the ModDrop method that drops modalities with a certain probability during training to break the dependency between modalities and improve model robustness to missing modalities. 
Xiao et al.~\cite{xiao2020audiovisual} claimed that dropping the modality with a faster learning pace could slow down its converging and facilitate the training of multimodal network. However, the previous modality dropout methods usually fix the drop probability of each modality during the whole training process~\cite{neverova2015moddrop,xiao2020audiovisual}, which can not well match the dynamic learning process of multiple modalities, especially in the case that modalities vary in learning pace. Hence, our OPM method adaptively adjusts the drop probability of each modality during training by monitoring the discriminative discrepancies between modalities, thereby focusing more on the less discriminative modalities.

\subsection{Generalization and stochastic gradient noise}
Based on the existing studies, the gradient noise in SGD is considered to have an essential correlation with the generalization ability of deep models~\cite{zhu2019anisotropic,chaudhari2018stochastic,xie2021artificial,wu2020noisy,he2019control}. This stochastic gradient noise brought by random mini-batch sampling, is believed that can serve as regularization and assist the deep model to escape from saddle point or local optimum~\cite{jastrzkebski2017three,chaudhari2018stochastic,xie2021artificial,wu2020noisy}. Jin et al.~\cite{jin2017escape} proposed that the suitable perturbation or noise in the gradient can help the model to escape saddle points efficiently. Neelakantan et al.~\cite{neelakantan2015adding} demonstrated that adding noise to gradient is helpful to potentially improve the training of deep neural networks. Zhou et al.~\cite{zhou2019toward} further provided theoretical proof that the stochastic gradient algorithms with proper Gaussian noise, are guaranteed to converge to the global optimum in polynomial time with random initialization. In our OGM method, to enhance the generalization ability of the multimodal model, we introduce extra Gaussian noise into the modified gradient.

\section{Method and analysis}

\subsection{Imbalanced learning analysis}
\label{sec:analysis}
As demonstrated in~\autoref{fig-teaser}, the uni-modal encoders in the jointly trained multimodal model are under-optimized to a different degree, and some modalities are worse learnt than others. In this section, we analyze this imbalanced phenomenon and find that the modality with more discriminative information dominates the optimization progress of multimodal model, causing other modalities to be worse under-optimized. In the analysis, we consider the widely-used late-fusion multimodal model. It should be noted that our proposed methods are also applicable to more complex cross-modal interactions in practice (as demonstrated by experiments in~\autoref{sec:multi-modal-task}), although our analysis focuses on the general late-fusion multimodal model.
Each modality is processed by the corresponding uni-modal encoder. Then their features are fused by the concatenation operation and passed to a single-layer linear classifier to produce the final prediction. The cross-entropy function is used as the discriminative learning objective.

For convenience, the dataset is denoted by $S=\{(x_{i}, y_{i})\}_{i=1,2...N}$. Suppose the number of modalities is $M$. Each $x_{i}$ contains inputs of $M$ modalities: $x_{i}=(x^{1}_{i}, x^{2}_{i}, \cdots, x^{M}_{i})$. $y_{i} \in \{1,2,\cdots,C\}$ is the target label of sample $x_i$ and $C$ is the number of categories. For modality $m$, where $m \in \{1,2,\cdots,M\}$, its input is processed by the corresponding encoder $\varphi^{m}(\theta^{m},\cdot)$. $\theta^{m}$ are the parameters of encoder. After extraction, their features are fused via concatenation, and passed to a single-layer linear classifier. $W \in \mathbb{R}^{C \times \sum^M_{m=1} d_{\varphi^m}}$ and $b \in \mathbb{R}^{C}$ denote the parameters of the linear classifier. $d_{\varphi^m}$ is the output dimension of $\varphi^{m}(\theta^{m},\cdot)$.

\subsubsection{Feed-forward stage} 
\label{sec:forward-stage}
With the above notions, the logits output of the considered multimodal model can be formulated as follows:
\begin{equation}
\label{output}
f(x_{i})=W[\varphi^{1}(\theta^{1},x^{1}_{i});\varphi^{2}(\theta^{2},x^{2}_{i});\cdots;\varphi^{M}(\theta^{M},x^{M}_{i})]+b.
\end{equation}

To observe the uni-modal components individually, we can mathematically transform the calculation of output $f(x_i) \in  \mathbb{R}^{C}$, and then~\autoref{output} is rewritten as:
\begin{equation}
\label{output_re}
f(x_{i})=W^{1}\cdot\varphi^{1}_i+W^{2}\cdot\varphi^{2}_i+\cdots+ W^{M}\cdot\varphi^{M}_i+b,
\end{equation}
where $\varphi^{m}(\theta^{m},x^{m}_{i})$ is denoted as $\varphi^{m}_i$ for simplicity. $W$ is divided into $M$ blocks: $[W^{1};W^{2};\cdots; W^{M}]$. $W^{m}\in \mathbb{R}^{C \times d_{\varphi^m}}$. Based on~\autoref{output_re}, the prediction of multimodal model is determined by \emph{the sum of uni-modal components}, $W^{m}\cdot\varphi^{m}_i$, in the feed-forward stage.

\subsubsection{Back-propagation stage}
Here we consider the cross-entropy loss function and the \emph{Gradient Descent} (GD) optimization method. The loss of sample $x_i$ is $\ell(x_i,y_i)= - \text{log} \frac{e^{f(x_{i})_{y_i}}}{\sum^C_{c=1}e^{f(x_{i})_{c}}}$, where $C$ is the number of categories, and $f(x_{i})_{c}$ is the logits for class $c$. During optimization, for modality $m$, where $m \in \{1,2,\cdots,M\}$, $W^{m}$ and $\varphi^{m}(\theta^{m},\cdot)$ are updated as:
\begin{equation}
\label{update_W}
\begin{aligned}
W_{t+1}^{m} &=W_{t}^{m}-\eta  \frac{1}{N} \sum^{N}_{i=1} \nabla_{W^{m}} \ell(x_i,y_i)\\
&=W_{t}^{m}-\eta \frac{1}{N} \sum^{N}_{i=1} \frac{\partial \ell(x_i,y_i) }{\partial f(x_{i})} \varphi^{m}_{i,t},
\end{aligned}
\end{equation}
\begin{equation}
\label{update_phi}
\begin{aligned}
\theta_{t+1}^{m} &=\theta_{t}^{m}-\eta \frac{1}{N} \sum^{N}_{i=1} \nabla_{\theta^{m}} \ell(x_i,y_i)\\
&=\theta_{t}^{m}-\eta  \frac{1}{N} \sum^{N}_{i=1} \frac{\partial \ell(x_i,y_i) }{\partial f(x_{i})}  \frac{\partial( W^{m}_{t}\cdot\varphi^{m}_{i,t})}{\partial \theta_{t}^{m}},
\end{aligned}
\end{equation}
where $\eta$ is the learning rate. Referring to~\autoref{update_W} and~\autoref{update_phi}, the update of $W^m$ and parameters in $\varphi^{m}$ has no correlation with the other modality, except the term related to the loss, \emph{i.e.,} $\frac{\partial \ell(x_i,y_i) }{\partial f(x_{i})}$. The uni-modal encoders thus are hard to make adjustments according to the feedback from each other. Denote logit of class $c$ is denoted as $f(x_{i})_{c}$. Then, the gradient $\frac{\partial \ell (x_i,y_i)}{\partial f(x_{i})_c}$ for category $c$ can be written as:
\begin{equation}
\label{ce_bp}
\begin{aligned}
\frac{\partial \ell(x_i,y_i) }{\partial f(x_{i})_c}=
\frac{e^{(W^{1}\cdot\varphi^{1}_i+W^{2}\cdot\varphi^{2}_i+\cdots+ W^{M}\cdot\varphi^{M}_i+b)_{c}}}{\sum^C_{j=1}e^{(W^{1}\cdot\varphi^{1}_i+W^{2}\cdot\varphi^{2}_i+\cdots+ W^{M}\cdot\varphi^{M}_i+b)_{j}}} - 1_{c=y_{i}},
\end{aligned}
\end{equation}
According to~\autoref{ce_bp}, the term $1_{c=y_{i}}$ is constant and not related to specific modality. For the first term, for each category, its denominator is the same. And the value of its molecule is determined by the sum of uni-modal components, $W^{m}\cdot\varphi^{m}_i$, for category $c$. Therefore, the concrete value of gradient for category $c$, $\frac{\partial \ell (x_i,y_i) }{\partial f(x_{i})_c}$, is also controlled by \emph{the sum of uni-modal components}, although it is not analytically equal to the sum of corresponding uni-modal components.

\subsubsection{Dominated optimization process}
The recent study has empirically shown that the different modalities could vary in the optimization process~\cite{wang2020makes}.
Meanwhile, our analysis of the feed-forward and back-propagation stages shows that both the multimodal prediction and the gradient values are controlled by \emph{the sum of uni-modal components}. Therefore, when one modality, such as modality $m$, has more discriminative information, it would dominate the multimodal prediction $f(x_{i})$ and gradient $\frac{\partial \ell (x_i,y_i)}{\partial f(x_{i})}$ via $W^{m}\cdot\varphi^{m}_i$. Even if another modality is under-optimized and yields incorrect results, the component from the better-performing modality $m$ can still ``correct'' these errors during summation, thus influencing both the feed-forward and back-propagation stages. Therefore, with \autoref{output_re} and \autoref{ce_bp},
another modality still with relatively lower confidence about the correct category, only earns limited optimization efforts, leading to it being underutilized. 
Overall, based on the above analysis, the modality with better performance dominates the optimization progress. Inevitably, as the multimodal model approaches convergence, the less discriminative modalities could still require further training due to their under-optimized features.

Additionally, the analysis in this section is based on the multimodal model that uses a single-layer classifier. For more general cases, we extend this analysis to the multimodal model that uses a multi-layer classifier with non-linear activation function in Appendix A.

\subsection{On-the-fly modulation strategies}
\label{sec:method}

\begin{figure*}[h]
    \centering
    \includegraphics[width=1\linewidth]{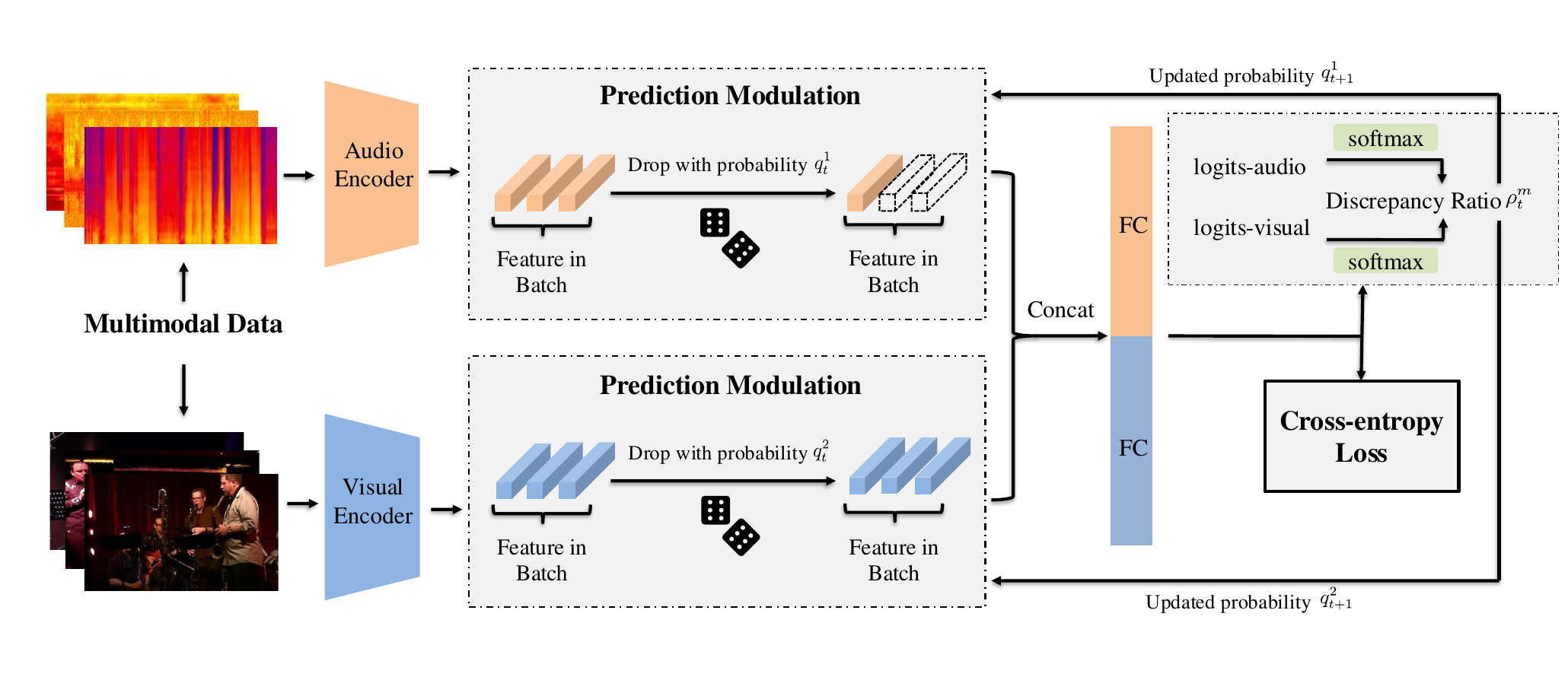}
    \vspace{-3em}
    \caption{\textbf{The pipeline of the On-the-fly Prediction Modulation.} Here we take two modalities as examples. In the feed-forward stage, the feature of modality $m$ is randomly dropped with probability $q^m$, where the probability is determined by the discriminative discrepancy ratio at the last iteration. Via OPM, the remained feature of suppressed modality could affect the multimodal prediction more, accordingly improving its learning.}
    \label{fig:OPM}
\end{figure*}

\begin{algorithm}[t]
\caption{Multimodal learning with OPM strategy}
\label{alg:opm}
\begin{algorithmic}
\Require Training dataset $S=\{(x_{i}, y_{i})\}_{i=1,2...N}$, iteration number $T$, initialized modal-specific parameters $\theta^{m}$, $m \in \{1,2,\cdots,M\}$.
\For{$t=0,\cdots,T-1$} 
    \State Sample a fresh mini-batch $B_{t}$ from $S$;
    \State Feed-forward the batched data $B_{t}$ to the model;
    \State Calculate $\rho^{m}_t$ using~\autoref{equ:uni-contribution} and~\autoref{equ:rho};
    \State Calculate $q_{t+1}^{m}$ using~\autoref{equ:drop};
    \State Drop the input of each modality with probability $q_{t}^{m}$;
    \State Calculate gradient using back-propagation;
    \State Update the model parameters.
\EndFor
\end{algorithmic}
\end{algorithm}

\subsubsection{On-the-fly prediction modulation}
\label{method-opm}
In the multimodal dataset, there often exists a dominant modality with more discriminative information. 
The overall performance of the model tends to be more dependent on the modality with more discriminative information, which in turn influences the optimization of other modalities.
Hence, to weaken the reliance on the dominant modality, we propose randomly dropping the feature of the more discriminative modality with a specific probability during the feed-forward stage, thus specifically accelerating the training of the suppressed modality. 
What's more, the discriminative ability of uni-modal features is gradually improved but at a different rate during training. Hence, the discrepancy in the discriminative ability between uni-modal features is dynamic. Correspondingly, it is necessary to adaptively adjust the drop probability of each modality during the training. Overall, in the proposed OPM method, the drop probability of modality with more discriminative information is adaptively adjusted during training based on the discriminative discrepancy degree between modalities. The pipeline of our OPM method is shown in~\autoref{fig:OPM}.

Here we follow the notation in~\autoref{sec:analysis}. To monitor the discriminative discrepancy between modalities during training, we first propose to estimate the uni-modal discriminative performance via:
\begin{equation}
\label{equ:uni-contribution}
\begin{gathered}
s_{i}^{m}=  \sum_{c=1}^C 1_{c=y_{i}} \cdot  	\iota (W^{m}_{i} \cdot \varphi^{m}_{i}(\theta^{m},x_{i}^{m})+\frac{b}{M})_{c},\\
\end{gathered}
\end{equation}
where $	\iota$ is the softmax function and $y_i$ is the ground truth label of sample $x_i$. As stated in~\autoref{equ:uni-contribution}, for modality $m$, the uni-modal component in the final multimodal prediction, $(W^{m}_{i} \cdot \varphi^m_{i}(\theta^{m},x_{i}^{m})+\frac{b}{M})$, is used as its the approximated prediction. Here we split the bias term into $\frac{b}{M}$ to estimate uni-modal performance. Since the bias term often has less effect on the prediction, its split could not have a great influence on the estimation. In~\autoref{sec:bias}, we provided ablation studies about the split of bias term. 

Since the modality with more discriminative information tends to have higher confidence for the correct category, \emph{i.e.,} the value of $s$ tends to be higher, then we propose to measure the discriminative discrepancy ratio of modality $m$ to other modalities by:
\begin{equation}
\label{equ:rho}
\rho^{m}_{t}= \frac{1}{M-1} \sum_{j\in[M],j\neq m} \frac{\sum_{i \in B_{t}}  s_{i}^{m} } {\sum_{i \in B_{t}}  s^{j}_i}.
\end{equation}
When the average uni-modal discriminative performance ratio to other modalities is larger than $1$, \emph{i.e.,} $\rho^{m}_{t} >1$, modality $m$ is more discriminative. $B_{t}$ is a random mini-batch which is chosen in the $t$-th step.

With $\rho^m_t$ to dynamically monitor the discriminative discrepancy among modalities, we can adaptively adjust the drop probability of modality $m$ through:
\begin{equation}
\label{equ:drop}
q^{m}_{t+1}=\left\{\begin{array}{cl}
 q_{base} \cdot (1+\lambda \cdot z ( \rho^{m}_{t})) & \text { }\rho ^{m}_{t}>1 \\
0 & \text { otherwise, }
\end{array}\right.
\end{equation}
where the base drop probability is $q_{base}$ and $z(\cdot)$ is a monotonically increasing function with a value range between $0$ and $1$. Hyper-parameter $q_{base}$ ranging in $(0,1)$ controls the range of modality dropout probability and $\lambda>0$ determines the degree of adjustment. Based on the modulation of OPM, for modality with more discriminative information ($\rho^m_{t}>1$), the discrepancy degree, \emph{i.e.,} $\rho_{t}^m$, is processed by $z(\cdot)$ to map it into $(0,1)$ as the increase of base drop probability. The drop probability of less discriminative one ($\rho_{t}^m \leq 1$) is set to $0$.

\begin{figure*}[t]
    \centering
    \includegraphics[width=1\linewidth]{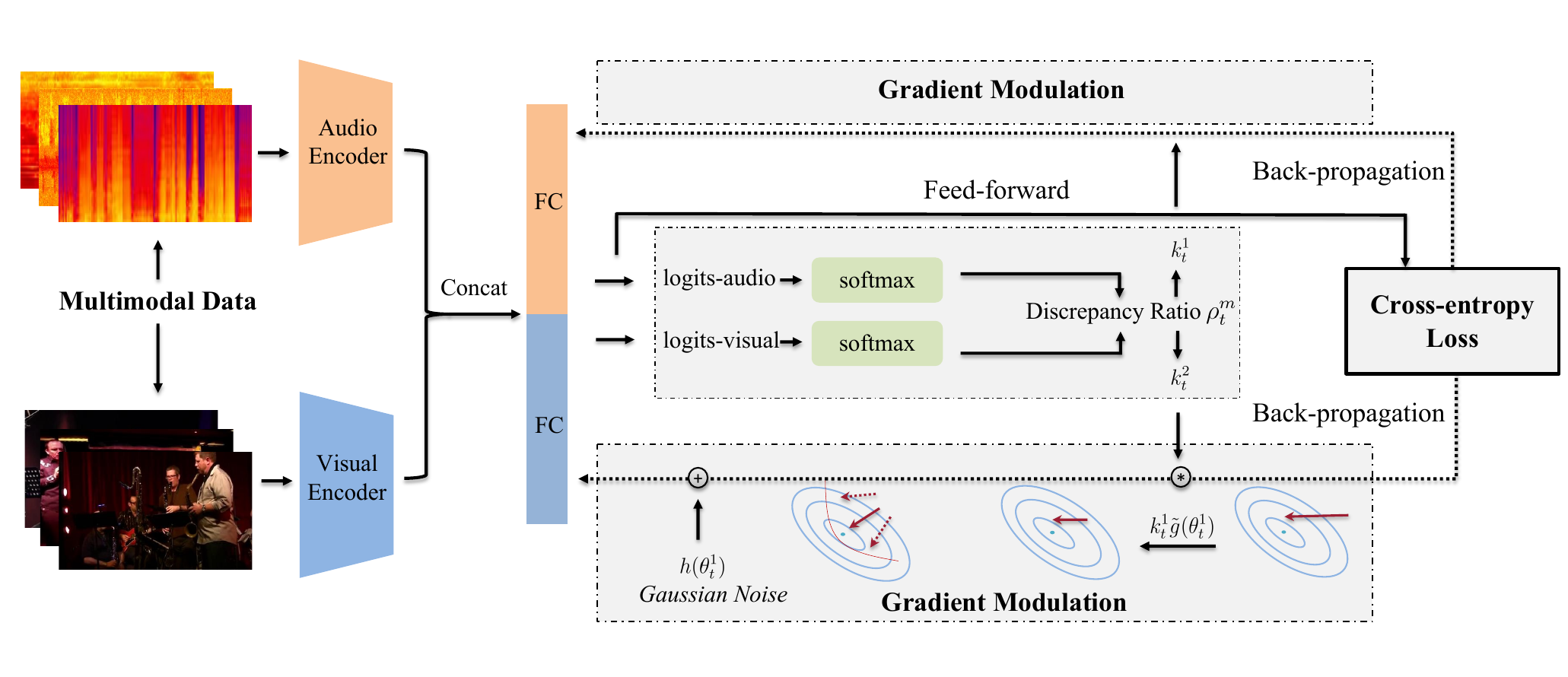}
    \vspace{-3em}
    \caption{\textbf{The pipeline of the On-the-fly Gradient Modulation strategy.} Here we take two modalities as example. In the back-propagation stage, the gradient of modality $m$ is modulated with $k^m$, which is determined by the discriminative discrepancy ratio at this iteration. Via OGM, the gradient of modality with more discriminative information is weakened, while the remained modality is not affected and can gain more training.}
    \label{fig:OGM}
\end{figure*}

\begin{algorithm}[h]
\caption{Multimodal learning with OGM strategy}
\label{alg:ogm}
\begin{algorithmic}
\Require Training dataset $S=\{(x_{i}, y_{i})\}_{i=1,2...N}$, iteration number $T$, initialized modal-specific parameters $\theta^{m}$, $m \in \{1,2,\cdots,M\}$.
\For{$t=0,\cdots,T-1$} 
    \State Sample a fresh mini-batch $B_{t}$ from $S$;
    \State Feed-forward the batched data $B_{t}$ to the model;
    \State Calculate $\rho^{m}_t$ using~\autoref{equ:uni-contribution} and~\autoref{equ:rho};
    \State Calculate $k_{t}^{m}$ using~\autoref{equ:ogm};
    \State Calculate gradient $\tilde{g}(\theta_{t}^{m})$ using back-propagation;
    \State Sample $h(\theta_{t}^{m})$ based on covariance of gradient $\tilde{g}(\theta_{t}^{m})$;
    \State Update using $\theta^{m}_{t+1} =\theta_{t}^{m} - \eta (k_{t}^{m}\tilde{g}(\theta_{t}^{m})+ h(\theta_{t}^{m}))$.
\EndFor
\end{algorithmic}
\end{algorithm}

When $\rho ^{m}_{t}>1$, the concrete value of drop probability $q_{t+1}^m$ ranges in $(q_{base},q_{base}\cdot(1+\lambda))$. In experiments, the maximum value of $q_{t+1}^m$ is set to $1$ to avoid illegal value. $tanh(x-1)$ function is used as $z(x)$\footnote{More alternative strategies and analysis are provided in~\autoref{sec:variants}.}. In addition, when the discrepancy between modalities is tiny, \emph{i.e.,} $\rho^{m}_{t}$ is close to $1$, the dropout probability $q_{t+1}^m$ is close to $q_{base}$, since $tanh(\rho^{m}_{t}-1)$ is approaching $0$.

The overall OPM method is provided in~\autoref{alg:opm}. The OPM method alleviates the imbalanced learning problem in the feed-forward stage. 
Via the proposed strategy, the effect on multimodal prediction of modality with better performance is relatively weakened. The larger the discriminative discrepancy, the stronger the modulation. Consequently, the previously suppressed modality could have a greater influence on multimodal prediction, thereby improving its learning. In addition, based on recent study~\cite{wei2020implicit}, the dropout strategy introduces additional noise into the gradient, which can improve model generalization. This is because gradient noise can help the model converge to wider minima, which typically generalize better. Therefore, the OPM method could also bring better multimodal generalization ability.

\subsubsection{On-the-fly gradient modulation}
\label{method-ogm}
The OPM method enhances the optimization of suppressed modality in the feed-forward stage. As discussed in~\autoref{sec:analysis}, the modality with more discriminative information also dominates the gradient during the back-propagation stage, leading to lower loss, and then limiting the gradient of other modalities.
Then, the OGM modulation strategy is proposed to amend the optimization of each modality in the back-propagation stage by mitigating the gradient of more discriminative modality based on the modality discrepancy during training. The pipeline of OGM is shown in~\autoref{fig:OGM}. 

Specifically, when using GD optimization method, for modality $m$, the parameters $\theta^{m}$ of the encoder $\varphi^{m}$ is updated as follows:
\begin{equation}
\label{theta_gd}
\theta_{t+1}^{m} =\theta_{t}^{m}-\eta \nabla_{\theta^{m}} \mathcal{L}(\theta_{t}^{m}),
\end{equation}
where $\nabla_{\theta^{m}} \mathcal{L}(\theta^{m}_t)=\frac{1}{N}\sum^N_{i=1}\nabla_{\theta^{m}} \ell (x_i;\theta^{m}_t)$ is the full gradient over all training samples. $\ell$ is the loss function.

In practice, we use the widely used \emph{Stochastic Gradient Descent} (SGD) optimization method, the parameters are updated as:
\begin{equation}
\label{theta_sgd}
\theta^{m}_{t+1} =\theta_{t}^{m}-\eta \tilde{g}(\theta_{t}^{m}),
\end{equation}
where $\tilde{g}(\theta_{t}^{m})=\frac{1}{|B_{t}|}\sum_{x\in  B_{t}}\nabla_{\theta^{m}} \ell (x;\theta_{t}^{m})$ is the gradient of current mini-batch $B_{t}$, and it can be considered as an unbiased estimation of the full gradient $\nabla_{\theta^{m}} \mathcal{L}(\theta^{m}_t)$~\cite{jastrzkebski2017three}. $|B_t|$ is the number of samples in this mini-batch.

Specific to the imbalanced learning problem, OGM proposes to control the uni-modal optimization in the back-propagation stage by modulating the gradient of each modality. Concretely, the gradient of modality with more discriminative information is mitigated, then the overall training process is slowed down and another modality could gain more optimization efforts. Similarly, considering the discriminate discrepancy is dynamic in the training process, the degree of gradient mitigation is adaptively adjusted. As in~\autoref{method-opm}, the discriminative discrepancy is also formulated as $\rho_t^m$ in~\autoref{equ:rho}. By means of $\rho_t^m$, we can dynamically modulate the gradient by:
\begin{equation}
\label{equ:ogm}
k^{m}_{t}=\left\{\begin{array}{cl} 1- \alpha \cdot z(\rho^{m}_{t}) & \text { }\rho ^{m}_{t}>1 \\
1 & \text { otherwise, }
\end{array}\right.
\end{equation}
where $\alpha>0$ is a hyper-parameter to control the degree of gradient modulation and $z(\cdot)$ is a monotonically increasing function with a value range between $0$ and $1$. Then, when $\rho ^{m}_{t}>1$, the concrete value of $k^{m}_{t}$ ranges in $(1-\alpha,1)$. In experiments, the minimum value of $k^{m}_{t}$ is set to $0$ to avoid illegal value. Also, $tanh(x-1)$ function is used as $z(x)$ in experiments.
After integrating the coefficient $k_t^m$, $\theta_{t}^{m}$ in iteration $t$ is updated as follows:
\begin{equation}
\label{equ:ogm-update}
\theta^{m}_{t+1} =\theta_{t}^{m}-\eta \cdot k_{t}^{m}\tilde{g}(\theta_{t}^{m}).
\end{equation}

After modulation, the gradient of modality with better performance ($\rho ^{m}_{t}>1$) is mitigated, and the other modality is not affected. The larger the discriminative discrepancy, the stronger the modulation. Correspondingly, the overall training process is slowed down and the suppressed modality can gain more training, easing the imbalanced learning problem. In addition, when the discrepancy between modalities is tiny, \emph{i.e.,} $\rho^{m}_{t}$ is close to $1$, the gradient modulation coefficient $k^{m}_{t}$ is also close to $1$, since $tanh(\rho^{m}_{t}-1)$ is approaching $0$. In this case, the gradient of relatively better modality is only slightly mitigated.

However, since the gradient mitigation operation could potentially harm the model generalization ability as the following analysis~\cite{jastrzkebski2017three}, we further introduce additional noise into the gradient to enhance generalization.

Concretely, as stated before, gradient of mini-batch $B_t$, $\tilde{g}(\theta_{t}^{m})$, is an un-biased estimation of full gradient $\nabla_{\theta^{m}} \mathcal{L}(\theta_{t}^{m})$. For a sufficiently large batch size, based on the central limit theorem, the gradient of each mini-batch is approximately Gaussian~\cite{jastrzkebski2017three}:
\begin{equation}
\tilde{g}(\theta^{m}_t)  \sim \mathcal{N}(\nabla_{\theta^{m}} \mathcal{L}(\theta^{m}_t), \frac{U}{|B_t|}),
\end{equation}
where $U$ is the covariance matrix, brought by the random sampling of SGD. Then the parameter update of SGD can be rewritten as follows:
\begin{equation}
\label{equ:noise}
\begin{gathered}
\theta^{m}_{t+1} =\theta_{t}^{m}-\eta \tilde{g}(\theta_{t}^{m}), \\
\theta_{t+1}^{m} = \theta_{t}^{m}-\eta \nabla_{\theta^{m}} \mathcal{L}(\theta^{m}_t)+ \xi_{t}, \xi_{t} \sim  \mathcal{N}(0,  \frac{\eta^2}{|B_t|}U),
\end{gathered}
\end{equation}
where $\xi_{t}$ is considered as the SGD noise term. Based on the existing studies~\cite{jastrzkebski2017three,zhu2019anisotropic}, this noise term in SGD has a close relationship with model generalization ability. The larger SGD noise often tends to bring better generalization, since it could help the model converge at a wider minima. Based on~\autoref{equ:noise}, we can have that the strength of SGD noise term $\xi_{t}$ is proportional to the ratio of learning rate to batch size. To verify that larger noise has better generalization, we conduct experiments with different $|B_t|$ and $\eta$. As shown in~\autoref{tab:bs_lr}, larger $\eta$ or smaller $|B_t|$ indeed bring better multimodal model performance.

With the above analysis, in OGM method, when modulating the gradient via coefficient $k_{t}^{m}$, $\theta^{m}$ is updated as:
\begin{equation}
\label{equ:noise-harm}
\begin{gathered}
\theta^{m}_{t+1} =\theta_{t}^{m}-\eta \cdot k_{t}^{m}\tilde{g}(\theta_{t}^{m}), \\
\theta_{t+1}^{m} =\theta_{t}^{m}-\eta  \cdot k_{t}^{m} \nabla_{\theta^{m}} \mathcal{L}(\theta^{m}_t)  + \xi_{t}^{\prime}, \\
\xi_{t}^{\prime} \sim  \mathcal{N}(0, (k_{t}^{m})^{2}\cdot \frac{\eta^2}{|B_t|}U).
\end{gathered}
\end{equation}

According to~\autoref{equ:ogm}, when modality $m$ is more discriminative with $\rho^m_t>1$, the modulation coefficient $k_t^m<1$. Inevitably, when the learning rate and batch size are fixed, the modulated SGD noise term $\xi_{t}^{\prime}$ is smaller than the original $\xi_{t}$, leading to the generalization ability of multimodal model could be affected. Therefore, it is desirable to recover the generalization ability. 

To avoid this potential generalization reduction issue, we propose to introduce a randomly sampled Gaussian noise $h(\theta_{t}^{m})  \sim  \mathcal{N}(0,\frac{U}{|B_t|})$ into the gradient to recover the strength of noise term:
\begin{equation}
\label{equ:ge}
\begin{aligned}
\theta^{m}_{t+1} &=\theta_{t}^{m} - \eta \cdot (k_{t}^{m}\tilde{g}(\theta_{t}^{m})+ h(\theta_{t}^{m}))\\
&=\theta_{t}^{m}-\eta  \cdot k_{t}^{m} \nabla_{\theta^{m}} \mathcal{L}(\theta^{m}_t) + \xi_{t}^{\prime}+  \epsilon _{t}.
\end{aligned}
\end{equation}

The additional Gaussian noise $h(\theta_{t}^{m})$ has a zero mean and the same covariance as current $\tilde{g}(\theta_{t}^{m})$ at iteration $t$. Then, $\epsilon _{t} \sim  \mathcal{N}(0,\frac{\eta^2}{|B_t|}U)$. Since $\epsilon_{t} $ and $\xi_{t}^{\prime} $ are two independent variants, and can be combined as a single term, then~\autoref{equ:ge} can be rewritten as:
\begin{equation}
\label{equ:ogm-final}
\begin{gathered}
\theta^{m}_{t+1} =\theta_{t}^{m}-\eta  \cdot k_{t}^{m} \nabla_{\theta^{m}} \mathcal{L}(\theta^{m}_t)+ \xi_{t}^{\prime\prime},\\
\xi_{t}^{\prime\prime} \sim  \mathcal{N} (0, ((k_{t}^{m})^{2}+1) \cdot \frac{\eta^2}{|B_t|}U).
\end{gathered}
\end{equation}

Hence, the SGD noise term $\xi_{t}^{\prime\prime}$ is recovered and even enhanced, compared with the original $\xi_{t}$, avoiding the risk of harming multimodal model generalization. The overall OGM method is provided in~\autoref{alg:ogm}. The OGM method eases the imbalanced learning problem in the back-propagation stage. Using the proposed strategy, the gradient of modality with better performance is weakened with guaranteed generalization ability based on the discriminative discrepancy, providing the suppressed modality with more training.

\subsubsection{Model performance analysis}
Using our on-the-fly modulation methods, we ease the imbalanced multimodal learning problem by enhancing uni-modal learning. Then, our methods improve the quality of uni-modal features. Since the multimodal representation is the fusion of uni-modal features, the improvement of uni-modal feature quality is expected to bring the improvement of multimodal latent representation. In Appendix B, we provide an analysis of how this improvement in representation attribute to better multimodal performance. Both quantitative and qualitative analysis are provided to verify how our methods improve the multimodal representation.

\begin{table}[t]
\centering
\tabcolsep=3mm
\caption{Experiments about different batch size $|B_t|$ and learning rate $\eta$ on CREMA-D and Kinetics-Sounds dataset.}
\vspace{-1em}
\label{tab:bs_lr}
\begin{tabular}{c|cc|cc}
\toprule
\multirow{2}{*}{\textbf{Method}} & \multicolumn{2}{c|}{\textbf{CREMA-D}} & \multicolumn{2}{c}{\textbf{Kinetics-Sounds}} \\
                                 & \textbf{Acc}      & \textbf{mAP}      & \textbf{Acc}          & \textbf{mAP}         \\ \midrule[0.7pt]
$|B_t|=32$, $\eta=1e-5$                   & 40.5              & 48.5              & 47.6                  & 51.4                 \\
$|B_t|=32$, $\eta=1e-4$                    & 63.4              & 69.1              & 61.9                  & 65.7                 \\
$|B_t|=32$, $\eta=1e-3$                    & \textbf{66.9}     & \textbf{72.1}     & \textbf{63.0}         & \textbf{67.3}        \\
$|B_t|=64$, $\eta=1e-3$                   & 64.7              & 71.4              & 62.1                  & 66.4                 \\
$|B_t|=128$, $\eta=1e-3$                   & 61.9              & 66.9              & 58.7                  & 63.2                 \\ \bottomrule
\end{tabular}
\end{table}

\begin{table*}[t]
	\centering
        \caption{Combined with different fusion methods. Encoders of UCF-101 dataset are pre-trained on ImageNet. OF denotes Optical Flow modality. Audio/RGB-only and Visual/OF-only methods are individually trained uni-modal model.}
        \vspace{-1em}
        \setlength{\tabcolsep}{4mm}{
	\begin{tabular}{c|cc|cc|cc|cc}
		\toprule
\multirow{3}{*}{\textbf{Method}} & \multicolumn{2}{c|}{\textbf{CREMA-D}} & \multicolumn{2}{c|}{\textbf{Kinetics-Sounds}}    & \multicolumn{2}{c|}{\textbf{UCF-101}}             & \multicolumn{2}{c}{\textbf{VGGSound}} \\
&\multicolumn{2}{c|}{\textbf{(Audio+Visual)}}&\multicolumn{2}{c|}{\textbf{(Audio+Visual)}}&\multicolumn{2}{c|}{\textbf{(RGB+Optical Flow)}}&\multicolumn{2}{c}{\textbf{(Audio+Visual)}}\\
    & \textbf{Acc}      & \textbf{mAP}      & \textbf{Acc} & \multicolumn{1}{c|}{\textbf{mAP}} & \textbf{Acc} & \multicolumn{1}{c|}{\textbf{mAP}} & \textbf{Acc}      & \textbf{mAP} \\
        \midrule[0.7pt]
        Audio/RGB-only & 61.4 & 67.7 &  49.9 & 51.3 & 77.2 & 82.5 & 47.5 & 49.3  \\
        Visual/OF-only & 50.9 & 52.2 & 46.7& 48.1 & 59.9 & 64.1 &  28.7 & 28.3\\
        \midrule
		Concatenation & 66.9 &  72.1  & 63.0 & 67.3 & 80.5 & 86.4  & 52.7 & 54.9 \\
	    Concatenation-OPM & \textbf{75.1} & 81.2  &  \textbf{67.0} & \textbf{72.5} &  \textbf{81.9} & \textbf{88.0} & $\textbf{54.1}$ & \textbf{56.5} \\
		Concatenation-OGM & 74.6 &  \textbf{81.4}  & 65.4 & 71.4 & 81.5 & 87.6 & 53.6 &   55.9 \\
		\midrule
		Summation & 60.8 &  65.8  & 62.7 & 66.0 & 80.9 & 86.8 & 52.5 & 54.7 \\
		Summation-OPM & \textbf{63.4} & \textbf{70.7} & \textbf{66.5} & \textbf{72.3} & \textbf{81.9} &  \textbf{88.0} & \textbf{53.8} & \textbf{56.2} \\
		Summation-OGM & \textbf{63.4} & 69.8 & 65.4 & 70.8 & 81.3 & 87.6 & 52.9 & 54.9 \\
        \midrule
        FiLM~\cite{perez2018film} & 61.8 & 66.6  & 60.6 & 64.4 & 75.2 & 81.0 & 49.8 & 51.6 \\
        FiLM-OPM & 62.4 & \textbf{69.4}  & 64.5 & 68.8 & \textbf{75.7} & 81.3 &  \textbf{51.3} & \textbf{53.5} \\
        FiLM-OGM & \textbf{63.0} & 68.3 & \textbf{65.3} & \textbf{70.9} & \textbf{75.7} & \textbf{81.6} & 50.0 & 51.3   \\ 
		\bottomrule
	\end{tabular}}
 \vspace{-1em}
	\label{tab:fusion}
\end{table*}

\section{Experiment}

\subsection{Dataset}
\label{sec:dataset}

\noindent \textbf{CREMA-D~\cite{cao2014crema}} is an emotion recognition dataset with two modalities: audio and visual. This dataset contains 7,442 video clips of 2-3 seconds for 91 persons speaking short works with different emotions. It covers 6 most common emotions. In the experiments, all samples are randomly divided into a 6,698-sample training and validation set and a 744-sample testing set. 

\noindent \textbf{Kinetics-Sounds}~\cite{arandjelovic2017look} is an action recognition dataset with two modalities, audio and visual. It contains 31 human action classes selected from Kinetics dataset~\cite{kay2017kinetics}. All videos are manually annotated utilizing Mechanical Turk and cropped to 10 seconds long around the action. This dataset contains 19k 10-second video clips. In our experiments, we follow the original dataset division.

\noindent \textbf{UCF-101}~\cite{soomro2012UCF101} is an action recognition dataset with two modalities, RGB and optical flow. It has 13,320 videos from 101 action categories. UCF-101 dataset contains two modalities: RGB and optical flow. The entire dataset is divided into a 9,537-sample training set and a 3,783-sample test set according to the original setting. \textbf{UCF-101-Three} dataset introduces the additional RGB-Difference modality based on the UCF-101 dataset

\noindent \textbf{VGGSound}\cite{chen2020vggsound} is an event recognition dataset with two modalities, audio and visual. It contains over 200k clips for 309 different classes, covering a wide range of daily audio events. Each video has a duration of 10 seconds. In our experiment, the dataset division is the same as~\cite{chen2020vggsound}. 168,618 videos are used for training and validation, and 13,954 videos are used for testing.

\noindent \textbf{CMU-MOSI}~\cite{zadeh2016mosi} is a sentiment analysis dataset with three modalities, audio, video and text. It is annotated with utterance-level sentiment labels. In experiments, labels are binary to classify whether the sentiment is positive or negative. This dataset consists of 93 movie review videos segmented into 2,199 utterances. The division of dataset follows the official split.

\noindent \textbf{AVE}\cite{tian2018audio} is a video dataset for the audio-visual event localization task. It contains 4,143 10-second videos from 28 event categories. Each video consists of at least one 2-second long audio-visual event, covering a wide range of domains, including human activities, animal activities, music performances, and vehicle sounds. In our experiments, the division of the dataset follows the official split.

\noindent \textbf{MUSIC-AVQA}~\cite{li2022learning} is designed for audio-visual question answering under musical scenario. It contains 9,288 videos covering 22 instruments, with a total duration of over 150 hours and 45,867 Question-Answering pairs. Each video contains around 5 QA pairs on average. In experiments, we follow the official split for training, evaluation, and test sets.

\subsection{Experimental settings}
\label{sec:setting}
In our experiments, when not specified, ResNet-18~\cite{he2016deep} is used as the backbone. Concretely, for the visual encoder, we take multiple frames as the input, and feed them into the 2D network like~\cite{zhao2018sound} does; for the audio encoder, we modified the input channel of ResNet-18 from three to one like~\cite{chen2020vggsound} does and the rest parts remain unchanged; for the optical flow encoder, we stack the horizontal vector and vertical vector as one frame, then multiple frames are also put into the 2D network as~\cite{zhao2018sound} does. In addition, the encoders used for UCF-101 and UCF-101-Three are ImageNet pre-trained and encoders of other datasets are trained from scratch. For the CMU-MOSI dataset, transformer-based networks are used as the backbone and trained from scratch.

Videos in Kinetics-Sounds, UCF-101, UCF-101-Three and VGGSound datasets are extracted at 1fps and three frames are uniformly sampled as the visual input. Videos in the AVE dataset are extracted frames with 1fps and all ten frames are used as the visual input. The audio data is first re-sampled into 16KHz and transformed into a spectrogram with size $257\times1,004$ using a window with length of 512 and overlap of 353. For the CREMA-D dataset with only 2-3 seconds video, One visual frame is randomly extracted from each clip and the audio data is processed into a spectrogram of size $257\times299$ with a window length of 512 and overlap of 353. 

During training, we use SGD with momentum (0.9) as the optimizer. For CREMA-D, Kinetics-Sounds, UCF-101, UCF-101-Three, and VGGSound datasets, the learning rate is $1e-3$, and the batch size is $32$. For CMU-MOSI, AVE, and MUSIC-AVQA datasets, the learning rate is $1e-4$, and batch size is $64$. All models are trained on 2 NVIDIA RTX 3090 (Ti).

\noindent\textbf{Evaluation metric.} For multi-class classification tasks, the widely-used \textbf{Accuracy} and \textbf{mAP} are used as evaluation metric:
\begin{equation}
    \text{Acc}=\frac{\sum^A_{i=1} 1_{\hat{y_i}=y_{i}}}{A},
\end{equation}
where $A$ is the number of testing samples, and $\hat{y_i}$ is the prediction of model for sample $x_i$ in testing set.
\begin{equation}
\begin{gathered}
\text{AP}_c = \sum_n (R_{c,n} - R_{c,n-1}) P_{c,n}, \\
\text{mAP}= \frac{1}{C} \sum^C_{c=1} \text{AP}_c,
\end{gathered}
\end{equation}
where $C$ is the category number. For class $c$, $\text{AP}_c$ summarizes a precision-recall curve as the weighted mean of precisions achieved at each threshold, with the increase in recall from the previous threshold used as the weight. $P_{c,n}$ and $R_{c,n}$ are the precision and recall for class $c$ at the $n-$th threshold. For the binary classification task, sentiment analysis, \textbf{Accuracy} and \textbf{F1 score} are used as evaluation metric:
\begin{equation}
    \text{F1 score}= 2 \cdot \frac{Precision \cdot Recall} {Precision + Recall}.
\end{equation}
In practice, the widely-used Python package scikit-learn is used for evaluation.

\begin{table}[t]
 \vspace{-1em}
\centering
\tabcolsep=4mm
 \caption{Combined with cross-modal interaction methods on the Kinetics-Sounds and UCF-101 dataset.}
 \vspace{-1em}
	\begin{tabular}{c|cc|cc}
		\toprule
\multirow{2}{*}{\textbf{Method}} & \multicolumn{2}{c|}{\textbf{Kinetics-Sounds}} & \multicolumn{2}{c}{\textbf{UCF-101}} \\
    & \textbf{Acc}      & \textbf{mAP}      & \textbf{Acc} & \multicolumn{1}{c}{\textbf{mAP}}              \\
		\midrule[0.7pt]
            CentralNet\cite{vielzeuf2018centralnet} & 66.4 & 70.6  & 82.0 & 87.4  \\
            CentralNet-OPM & 68.3 & 73.5 & 82.9 & 88.1 \\
            CentralNet-OGM &  \textbf{68.8} & \textbf{73.9} & \textbf{83.2} & \textbf{88.8} \\
            \midrule
            VATT\cite{akbari2021vatt} & 61.8 &  64.0 & 81.2 & 86.5 \\
            VATT-OPM & \textbf{69.2} & \textbf{76.0} & \textbf{82.6} & \textbf{89.2} \\
            VATT-OGM & 67.6 & 73.3 & 82.3 & 88.0 \\
            \midrule
            MMTM\cite{joze2020mmtm} & 63.8 & 68.4 & 79.8 & 85.3 \\
            MMTM-OPM & \textbf{67.9} & \textbf{73.8} & \textbf{80.5} & \textbf{86.7} \\
            MMTM-OGM &  66.1 & 71.5 & 79.9 & 85.8\\
            \bottomrule
	\end{tabular}
 \vspace{-1em}
	\label{tab:inter}
\end{table}

\subsection{Comparison on the multimodal task}
\label{sec:multi-modal-task}
\subsubsection{Combination with different fusion methods}

We first apply OPM and OGM methods to two vanilla fusion methods: Concatenation and Summation. Additionally, we compared the specifically designed fusion method FiLM~\cite{perez2018film}. The results on four datasets are as shown in~\autoref{tab:fusion}. We also provide the performance of individually trained uni-modal  models for comparison. It can be observed that the uni-modal performance is imbalanced across different datasets, demonstrating that the discriminative ability of different modalities varies on different datasets. For instance, the performance of audio-only model outperforms the visual-only model on the sound-oriented VGGSound dataset. Furthermore, in some cases, the jointly trained multimodal model can be inferior to the best performing uni-modal models. For example, the performance of the audio-only model on CREMA-D dataset is better than the multimodal model with Summation fusion.

Moreover, OPM and OGM can bring considerable improvement to both vanilla and specifically designed fusion methods, which demonstrates the effectiveness and satisfactory flexibility of our strategies. As shown in~\autoref{fig:training_loss} and ~\autoref{tab:training_cost}, although our modulation for dominant modality slows down the overall training process a bit, our OPM and OGM methods do not bring much additional training time with the same training epoch, compared with Concatenation baseline. Additionally, it can be observed that the convergence loss of our OPM and OGM methods is higher than that of

\begin{minipage}[t]{\columnwidth}
\centering
\begin{minipage}[h]{0.45\columnwidth}
\centering
\includegraphics[height=0.8\columnwidth]{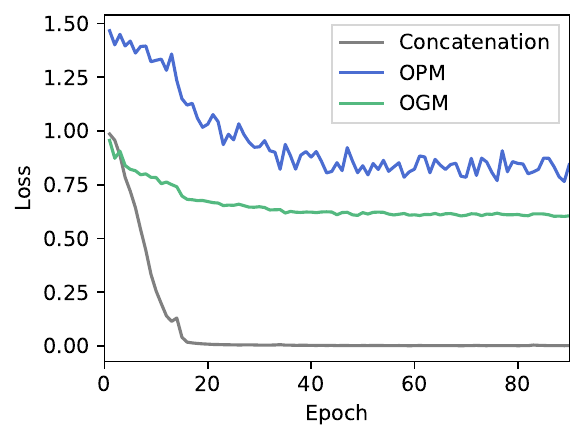}
\vspace{-2em}
\makeatletter\def\@captype{figure}\makeatother\caption{Training loss change on CREMA-D dataset.}
\label{fig:training_loss}
\end{minipage}
\quad
\quad
\begin{minipage}[h]{0.42\columnwidth}
\makeatletter\def\@captype{table}\makeatother\caption{Training time on CREMA-D dataset.}
\vspace{-1em}
\scalebox{0.8}{
\begin{tabular}{c|c}
\toprule
\textbf{Method} & \textbf{\begin{tabular}[c]{@{}c@{}}Training \\ time (Min)\end{tabular}} \\  \toprule
Concatenation   & $1.00\times$                \\ \midrule
GBlending~\cite{wang2020makes}   & $1.50\times$               \\
Greedy~\cite{wu2022characterizing}          & $1.01\times$                   \\ \midrule
OPM             & $1.01\times$                  \\
OGM             & $1.01\times$                  \\ \bottomrule
\end{tabular}}
\label{tab:training_cost}   
\end{minipage}
\end{minipage}
\vspace{1em}

\noindent the baseline, possibly because our modulation prevents the multimodal model from over-memorizing the training samples.

In most cases, the OPM method shows a more obvious enhancement than the OGM method. Based on the specific modulation design, the OPM method directly drops the feature of the better-performing modality in the feed-forward stage, allowing the suppressed modality to fully determine the multimodal prediction and temporarily control the optimization process. OGM weakens the gradient of the dominant modality in the back-propagation stage, slowing down the overall training. The modulation by OPM is stronger, which allows the suppressed modality to receive more targeted optimization efforts.

In addition, our methods show a more significant improvement on CREMA-D dataset compared to other datasets. The reason could be that data samples of CREMA-D dataset are recorded videos in controlled environments with less noise, while samples of other datasets are more noisy ``in the wild'' videos. Hence, for CREAM-D dataset, these clean samples are supposed to be easier to learn but suppressed by imbalanced multimodal learning. Then, after our modulation, multimodal model is more effectively enhanced.

\begin{table}[t]
\centering
\tabcolsep=2mm
\caption{Comparison with other modulation methods on CREMA-D, Kinetics-Sounds, and UCF-101 dataset. OPM and OGM methods are based on Concatenation fusion.}
\vspace{-1em}
\label{tab:modulation}
\begin{tabular}{c|cc|cc|cc}
\toprule
\multirow{2}{*}{\textbf{Method}} & \multicolumn{2}{c|}{\textbf{CREMA-D}} & \multicolumn{2}{c|}{\textbf{Kinetics Sounds}} & \multicolumn{2}{c}{\textbf{UCF-101}} \\
                                 & \textbf{Acc}      & \textbf{mAP}      & \textbf{Acc}          & \textbf{mAP}          & \textbf{Acc}      & \textbf{mAP}     \\ \midrule[0.7pt]
Concatenation                    & 66.9              & 72.1              & 63.0                  & 67.3                  & 80.5              & 86.4             \\ \midrule
GBlending~\cite{wang2020makes}                    & 72.0              & 78.6              & \textbf{67.5}         & 72.1                  & 80.9              & 86.6             \\
Greedy~\cite{wu2022characterizing}                           & 68.4              & 76.8              & 65.2                  & 69.6                  & 80.9              & 86.9             \\  \midrule
OPM                              & \textbf{75.1}     & 81.2              & 67.0                  & \textbf{72.5}         & \textbf{81.9}     & \textbf{88.0}    \\
OGM                              & 74.6              & \textbf{81.4}     & 65.4                  & 71.4                  & 81.5              & 87.6             \\ \bottomrule
\end{tabular}
\end{table}

\subsubsection{Cross-modal interaction scenarios}
Beyond the above vanilla and specifically designed fusion methods, various cross-modal interaction modules are proposed to improve the integration of different modalities in multimodal learning. In this section, we combine our methods with several multimodal models with cross-modal interaction modules, CentralNet\cite{vielzeuf2018centralnet}, VATT~\cite{akbari2021vatt} and MMTM~\cite{joze2020mmtm},  to evaluate their effectiveness in more complex cross-modal interaction scenarios. CentralNet~\cite{vielzeuf2018centralnet} integrates the uni-modal feature of intermediate layers. VATT~\cite{akbari2021vatt} uses the cross-modal attention mechanism. MMTM~\cite{joze2020mmtm} activates the intermediate features of one modality with the guidance of others via the squeeze and excitation module, which can be regarded as self-attention on channels. Based on the results shown in~\autoref{tab:inter}, our methods are not limited to the vanilla multimodal integration cases, but can also enhance the performance of multimodal models with more complex cross-modal interaction. 

In addition, we also observe the change in discrepancy ratio during training of cross-modal interaction methods. According to~\autoref{fig:inter-ratio}, CentralNet~\cite{vielzeuf2018centralnet} method reduces the discriminative discrepancy between modalities, since it enhances the cross-modal interaction at the mid-level. Furthermore, our methods can be applied to these more complex scenarios, further reducing the discrepancy ratio and alleviating the imbalanced learning problem.

\subsubsection{Comparison with uni-modal modulation methods}
To address the issue that multimodal models often cannot jointly utilize all modalities effectively, uni-modal modulation methods have been proposed. To demonstrate the advantage of OPM and OGM, we make comparisons with other modulation methods: GBlending~\cite{wang2020makes} and Greedy~\cite{wu2022characterizing}. These methods either add uni-modal loss functions to the multimodal objective, controlling the training of individual modalities by loss weight~\cite{wang2020makes}, or enhance the optimization of other modalities by additional training step~\cite{wu2022characterizing}. To ensure the fairness of experiments, the same backbone network and training settings are used. The comparison results are shown in~\autoref{tab:modulation}. We can have the following observations:

Firstly, GBlending~\cite{wang2020makes} and Greedy~\cite{wu2022characterizing} achieve better results than the baseline model with Concatenation fusion, which proves the effectiveness of these methods in solving the imbalanced under-optimized problem in multimodal models. Secondly, both our OPM and OGM methods achieve improvement compared with the Concatenation baseline. In particular, the OPM method obtains the best performance on the CREMA-D and UCF-101 datasets, outperforming other modulation methods. These results indicate the advantage of our on-the-fly modulation methods.

Moreover, it should be noted that GBlending~\cite{wang2020makes} requires training additional uni-modal classifiers to provide a reference for modulation. Although it achieves the best accuracy on the Kinetics-Sounds dataset, it incurs higher training costs. As shown in~\autoref{tab:training_cost}, the training time of GBlending~\cite{wang2020makes} is considerably longer. Our methods are much more efficient while being effective.

\begin{table}[t]
\centering
\tabcolsep=1.3mm
\caption{Comparison with other dropout methods on CREMA-D, Kinetics-Sounds, and UCF-101 dataset. OPM and OGM methods are based on Concatenation fusion.}
\vspace{-1em}
\label{tab:dropout}
\begin{tabular}{c|cc|cc|cc}
\toprule
\multirow{2}{*}{\textbf{Method}} & \multicolumn{2}{c|}{\textbf{CREMA-D}} & \multicolumn{2}{c|}{\textbf{Kinetics-Sounds}} & \multicolumn{2}{c}{\textbf{UCF-101}} \\
                                 & \textbf{Acc}      & \textbf{mAP}      & \textbf{Acc}          & \textbf{mAP}          & \textbf{Acc}      & \textbf{mAP}     \\ \midrule[0.7pt]
Concatenation                    & 66.9              & 72.1              & 63.0                  & 67.3                  & 80.5              & 86.4             \\ \midrule
ModDrop~\cite{neverova2015moddrop}                          & 69.9              & 77.6              & 65.9                  & 71.8                  & 81.4              & 87.2             \\
Dropout~\cite{srivastava2014dropout}                          & 66.5              & 71.8              & 62.8                  & 66.1                  & 80.3              & 85.8             \\
Annealed dropout~\cite{rennie2014annealed}                 & 65.3              & 70.9              & 62.1                  & 65.2                  & 80.1              & 84.7             \\
Evolutional dropout~\cite{li2016improved}              & 65.9              & 72.0              & 62.4                  & 65.7                  & 80.4              & 86.1             \\
Curriculum dropout~\cite{morerio2017curriculum}               & 64.9              & 70.2              & 61.9                  & 64.9                  & 79.9              & 84.3             \\ \midrule
OPM                              & \textbf{75.1}     & 81.2              & \textbf{67.0}                  & \textbf{72.5}         & \textbf{81.9}     & \textbf{88.0}    \\
OGM                              & 74.6              & \textbf{81.4}     & 65.4                  & 71.4                  & 81.5              & 87.6             \\ \bottomrule
\end{tabular}
\end{table}

\subsubsection{Comparison with other dropout methods}
In this section, we compare our methods with modality dropout method: ModDrop~\cite{neverova2015moddrop}, and neuron dropout methods: Annealed dropout~\cite{rennie2014annealed}, Evolutional dropout~\cite{li2016improved} and Curriculum dropout~\cite{morerio2017curriculum}. ModDrop~\cite{neverova2015moddrop} drops modalities with a certain probability during training, easing the reliance of the multimodal model on any specific modality and improving model robustness. The results are shown in~\autoref{tab:dropout}. Based on the results, neuron dropout methods that drop neurons with a certain probability do not work well for the imbalanced multimodal learning problem because they focus mainly on neuron-level co-adaptation without considering modality-level discrepancies. In addition, although it is not designed for balanced multimodal learning, ModDrop~\cite{neverova2015moddrop} still has a gain in performance. As analyzed in~\autoref{sec:analysis}, the reason could be that the modality dropout provides each modality a chance to be optimized independently, getting rid of the suppression of others. Moreover, our methods are superior to these dropout methods, since they adaptively adjust the modulation of each modality during training by monitoring the discriminative discrepancy between modalities, considering the dynamical training process.

\begin{table}[t]	
\centering
\tabcolsep=1.3mm
\caption{Accuracy of audio-visual event localization methods on AVE dataset, and audio-visual question answering methods on MUSIC-AVQA dataset.}
\vspace{-1em}
\label{tab:complex}
\begin{tabular}{ccccc}
\toprule
\multicolumn{5}{c}{\textbf{\begin{tabular}[c]{@{}c@{}}Audio-visual Event Localization\\ (Audio+Visual)\end{tabular}}}                                                        \\ \midrule[0.7pt]
\multicolumn{1}{c|}{\textbf{Method}} & \multicolumn{1}{c|}{\textbf{w/o}} & \multicolumn{1}{c|}{\textbf{w/ OPM}} & \multicolumn{1}{c|}{\textbf{w/ OGM}} & \textbf{w/ OPM+OGM} \\ \midrule
\multicolumn{1}{c|}{AVGA~\cite{tian2018audio}}            & \multicolumn{1}{c|}{73.6}         & \multicolumn{1}{c|}{75.6}            & \multicolumn{1}{c|}{\textbf{76.4}}   & 76.2                \\
\multicolumn{1}{c|}{AVSDN~\cite{lin2019dual}}           & \multicolumn{1}{c|}{74.3}         & \multicolumn{1}{c|}{\textbf{75.1}}   & \multicolumn{1}{c|}{74.7}            & 74.8                \\
\multicolumn{1}{c|}{PSP~\cite{zhou2021positive} }             & \multicolumn{1}{c|}{74.4}         & \multicolumn{1}{c|}{\textbf{75.4}}   & \multicolumn{1}{c|}{75.3}            & 75.3                \\ \bottomrule \toprule
\multicolumn{5}{c}{\textbf{\begin{tabular}[c]{@{}c@{}}Audio-visual Question Answering\\ (Audio+Visual+Text)\end{tabular}}}                                                   \\ \midrule[0.7pt]
\multicolumn{1}{c|}{\textbf{Method}} & \multicolumn{1}{c|}{\textbf{w/o}} & \multicolumn{1}{c|}{\textbf{w/ OPM}} & \multicolumn{1}{c|}{\textbf{w/ OGM}} & \textbf{w/ OPM+OGM} \\ \midrule
\multicolumn{1}{c|}{AVSD~\cite{schwartz2019simple}}            & \multicolumn{1}{c|}{66.9}         & \multicolumn{1}{c|}{\textbf{67.3}}   & \multicolumn{1}{c|}{67.2}            & 67.1                \\
\multicolumn{1}{c|}{PanoAVQA~\cite{yun2021pano}}        & \multicolumn{1}{c|}{70.3}         & \multicolumn{1}{c|}{70.8}            & \multicolumn{1}{c|}{71.0}            & \textbf{71.1}       \\ \bottomrule
\end{tabular}
\vspace{-0.5em}
\end{table}

\subsubsection{Complex multimodal task scenarios} 
The above experiments are either action recognition or emotion recognition, which are both video-level classifications. To further evaluate the proposed OPM and OGM methods in more general cases, we employ them on two complex multimodal tasks: audio-visual event localization and audio-visual question answering. The audio-visual event localization task aims to temporally demarcate both audible and visible events from a video. Therefore, it is in fact a segment-wise classification task. For each segment, there are predicted event labels and ground truth event labels. Global segment-wise classification accuracy is used as the evaluation metric in these experiments. The audio-visual question answering task involves selecting the correct answer for a given question about visual objects, sounds, and their associations. The widely-used AVE and MUSIC-AVQA datasets are used for these two tasks. Audio-visual event localization methods, AVGA~\cite{tian2018audio}, AVSDN~\cite{lin2019dual} and PSP~\cite{zhou2021positive}, are compared. Audio-visual question answering methods, AVSD~\cite{schwartz2019simple} and PanoAVQA~\cite{yun2021pano} are also compared.

\begin{table}[t]
\centering
\tabcolsep=4mm
\caption{Accuracy of emotion recognition task-oriented methods on the CREMA-D dataset.}
\label{tab:emotion}
 \vspace{-1em}
	\begin{tabular}{c|c|c|c}
		\toprule
		\textbf{Method} & \textbf{w/o}  & \textbf{w/ OPM} & \textbf{w/ OGM} \\
		\midrule
I-vector~\cite{heracleous2019comprehensive}                         & \multicolumn{1}{c|}{67.4}                               & \textbf{75.2}                                       & 74.6                                                                         \\
X-vector~\cite{pappagari2020x}                           & \multicolumn{1}{c|}{69.2}                               & \textbf{74.7}                                       & 73.9                                                                         \\
MWTSM~\cite{ghaleb2019multimodal}                            & \multicolumn{1}{c|}{68.4}                               & \textbf{74.9}                                       & 74.1                                                                         \\ \bottomrule
\end{tabular}
\end{table}

\begin{table}[t]
\centering
\tabcolsep=4mm
\caption{Combined with action recognition task-oriented methods on the Kinetics-Sounds and UCF-101 dataset.}
	\begin{tabular}{c|cc|cc}
		\toprule
\multirow{2}{*}{\textbf{Method}} & \multicolumn{2}{c|}{\textbf{Kinetics-Sounds}} & \multicolumn{2}{c}{\textbf{UCF-101}} \\
    & \textbf{Acc}      & \textbf{mAP}      & \textbf{Acc} & \multicolumn{1}{c}{\textbf{mAP}}              \\
		\midrule[0.7pt]
            TSN\cite{wang2016temporal} & 65.0 & 69.4 & 80.9 & 86.2 \\
            TSN-OPM & \textbf{70.0} & \textbf{75.9} & \textbf{82.5} & \textbf{87.7} \\
            TSN-OGM & 67.7 & 73.3 & 81.8 & 87.1 \\
            \midrule
            TSM\cite{lin2019tsm} & 66.0 & 70.2 & 81.2 & 86.5 \\
            TSM-OPM & \textbf{69.9} & \textbf{75.8} & 82.3 & 87.5 \\
            TSM-OGM & 67.5 & 73.0 & \textbf{82.4} & \textbf{87.9} \\
            \bottomrule
	\end{tabular}
 \vspace{-0.5em}
	\label{tab:action}
\end{table}

\begin{table}[t]
\centering
\caption{Combined with multi-layer classifier cases on CREMA-D, Kinetics-Sounds and UCF-101 dataset.}
\label{tab:threelayer}
\setlength{\tabcolsep}{1mm}{
\begin{tabular}{c|cc|cc|cc}
\toprule
\multirow{2}{*}{\textbf{Method}} & \multicolumn{2}{c|}{\textbf{CREMA-D}} & \multicolumn{2}{c|}{\textbf{Kinetics-Sounds}} & \multicolumn{2}{c}{\textbf{UCF-101}} \\
                                 & \textbf{Acc}      & \textbf{mAP}      & \textbf{Acc}          & \textbf{mAP}          & \textbf{Acc}      & \textbf{mAP}     \\ \midrule[0.7pt]
Multi-layer classifier           & 67.7              & 73.2              & 63.7                  & 68.4                  & 80.3              & 85.2             \\ \midrule
Multi-layer classifier-OPM       & \textbf{71.9}              & \textbf{79.7}              & \textbf{66.3}                  & \textbf{71.7}                  & \textbf{81.5}              & \textbf{87.3}             \\
Multi-layer classifier-OGM       & 69.3              & 77.9              & 64.5                  & 70.3                  & 81.2              & 87.0             \\ \bottomrule
\end{tabular}}
\end{table}

The experiment results, as shown in~\autoref{tab:complex}, indicate that both OPM and OGM maintain their effectiveness and enhance performance in these more complex multimodal tasks. This demonstrates that the imbalanced learning phenomenon is not limited to video-level classification; it also affects more general multimodal cases. Moreover, to effectively capture the temporal correlations among modalities, both audio-visual event localization methods and audio-visual question answering methods often have more cross-modal interaction modules. Although this interaction design brings difficulty for the uni-modal performance estimation, our methods still obtain consistent performance gain. These experiments also indicate that our methods are effective on more complex multimodal tasks.

\subsubsection{Combination with task-oriented methods}
In this section, we combine OPM and OGM methods with several task-oriented methods to further evaluate their flexibility. Emotion recognition methods, I-vector~\cite{heracleous2019comprehensive}, X-vector~\cite{pappagari2020x} and MWTSM~\cite{ghaleb2019multimodal} are compared on CREMA-D dataset. Action recognition methods, TSN~\cite{wang2016temporal} and TSM~\cite{lin2019tsm} are compared on Kinetics-Sounds and UCF-101 dataset.  For all these methods, we utilize ResNet-18 as the backbone for both RGB and optical flow modalities and encoders are pre-trained on ImageNet on the UCF-101 dataset.
In~\autoref{tab:emotion} and~\autoref{tab:action}, we provide the experiment results. 
Both our OPM and OGM methods can be integrated with these specifically designed task-oriented models to further enhance performance. The experiments in combination with task-oriented methods further demonstrate the versatility of our OPM and OGM strategies on the more general jointly trained multimodal model.

\subsubsection{Combined with multi-layer classifier case}

In the analysis and methods section, we suppose that the final multimodal classifier is a single fully-connected layer without activation function, where the uni-modal discriminative performance could be estimated by splitting the weight of the final classifier. In fact, the core idea of our methods is to estimate the uni-modal prediction, and then use the estimated uni-modal prediction to evaluate the discriminative ability discrepancy between modalities. Therefore, if the uni-modal prediction can be estimated, our methods are applicable beyond single-layer classifier scenarios. In this section, we consider the case that the final classifier is multi-layer. In such case, to estimate the uni-modal discriminative performance score $s^m$ for modality $m$, we retain features of modality $m$ while setting features of other modalities to $0$, obtaining its uni-modal prediction. This zero-out strategy can be used for most multimodal cases. In similar studies~\cite{ghorbani2020neuron,hu2022shape}, this zero-out strategy has been widely used. For instance, Hu et al.~\cite{hu2022shape} use zero-padding to replace modality $x$'s feature to effectively simulate the multimodal prediction when modality $x$ is absent. Hence, it is reasonable to utilize the zero-out strategy to simulate other modalities being absent, and estimate the prediction of remained modality. Based on the results in~\autoref{tab:threelayer}, our methods maintain their efficacy. These results indicate that our idea of prediction modulation and gradient modulation could work well and extend to more complex classifier cases.

\begin{table}[t]
\centering
 \caption{Experiments results on CMU-MOSI and UCF-101-Three datasets. Encoders of UCF-101-Three are pre-trained on ImageNet.}
 \vspace{-1em}
 \label{tab:thee_mod}
\begin{tabular}{c|cc|cc}
\toprule
\multirow{2}{*}{\textbf{Method}} & \multicolumn{2}{c|}{\textbf{\begin{tabular}[c]{@{}c@{}}CMU-MOSI\\ (Audio+Visual+Text)\end{tabular}}} & \multicolumn{2}{c}{\textbf{\begin{tabular}[c]{@{}c@{}}UCF-101-Three\\ (RGB+OF+RGB Diff)\end{tabular}}} \\
                                 & \textbf{Acc}                                  & \textbf{F1 score}                                 & \textbf{Acc}                                    & \textbf{mAP}                                   \\ \midrule[0.7pt]
Concatenation                    & 75.9                                          & 76.0                                                 & 82.5                                            & 88.6                                           \\ \midrule
OPM                              & \textbf{77.6}                                 & \textbf{77.4}                                        & \textbf{83.2}                                   & 89.0                                           \\
OGM                              & 76.8                                          & 76.7                                                 & 82.8                                            & 88.6                                           \\
OPM+OGM                          & 77.1                                          & 76.7                                                 & 83.2                                            & \textbf{89.2}                                  \\ \bottomrule
\end{tabular}
\end{table}

\begin{table}[t]
\centering
\tabcolsep=3mm
\caption{Experiments about the combination of our OPM and OGM methods. All methods are based on Concatenation fusion.}
\vspace{-1em}
\label{tab:both}
\begin{tabular}{c|cc|cc}
\toprule
\multirow{2}{*}{\textbf{Method}} & \multicolumn{2}{c|}{\textbf{CREMA-D}} & \multicolumn{2}{c}{\textbf{Kinetics-Sounds}} \\
                                 & \textbf{Acc}       & \textbf{mAP}     & \textbf{Acc}          & \textbf{mAP}         \\ \midrule[0.7pt]
Concatenation                    & 66.9               & 72.1             & 63.0                  & 67.3                 \\ \midrule
OPM                              & 75.1               & 81.2             & 67.0                  & 72.5                 \\
OGM                              & 74.6               & 81.4             & 65.4                  & 71.4                 \\
OPM+OGM                          & \textbf{76.7}      & \textbf{83.6}    & \textbf{68.0}         & \textbf{74.7}        \\ \bottomrule \toprule
\multirow{2}{*}{\textbf{Method}} & \multicolumn{2}{c|}{\textbf{UCF-101}} & \multicolumn{2}{c}{\textbf{VGGSound}}        \\
                                 & \textbf{Method}    & \textbf{Acc}     & \textbf{mAP}          & \textbf{Acc}         \\ \midrule[0.7pt]
Concatenation                    & 80.5               & 86.4             & 52.7                  & 54.9                 \\ \midrule
OPM                              & \textbf{81.9}      & \textbf{88.0}    & \textbf{54.1}         & \textbf{56.5}        \\
OGM                              & 81.5               & 87.6             & 53.6                  & 55.9                 \\
OPM+OGM                          & 81.5               & 87.8             & 53.3                  & 56.3                 \\ \bottomrule
\end{tabular}
\end{table}

\subsubsection{More-than-two modality cases}
In this section, to further evaluate the effectiveness of our methods in scenarios involving more than two modalities, we conduct experiments on the CMU-MOSI and UCF-101-Three datasets. CMU-MOSI dataset covers three modalities, audio, visual and text. UCF-101-Three dataset has three modalities, RGB, optical flow, and RGB difference. These scenarios are more challenging than typical cases with only two modalities due to the more complex relationships among modalities. Based on the results in~\autoref{tab:thee_mod}, both OPM and OGM remain effective in these more complex multimodal scenarios. This demonstrates the scalability of our methods and their ability to adapt to diverse datasets.

\subsubsection{Combination of OPM and OGM methods} 
Our OPM and OGM methods are proposed to target the feed-forward stage and back-propagation stage, respectively. To explore whether these two methods can be combined, we conduct experiments on all used datasets. The results are shown in~\autoref{tab:complex},~\autoref{tab:thee_mod} and~\autoref{tab:both}.  It is observed that using both OPM and OGM methods can yield better results. But this improvement is not consistent across all datasets. The reason could be that the estimation of uni-modal discriminative performance discrepancy is likely to be influenced when simultaneously conducting modulation at the feed-forward stage and back-propagation stage. For example, when dropping modalities first at the feed-forward stage, the modality discrepancy is changed. Then, the gradient modulation based on the original modality discrepancy is affected. Hence, how to well combine the modulation at both stages is expected to be further explored.

\subsection{Fine-grained effectiveness analysis}
\label{sec:fine-analysis}

\begin{figure}[t]
    \centering
    \subfigure[]{
    \includegraphics[width=0.47\linewidth]{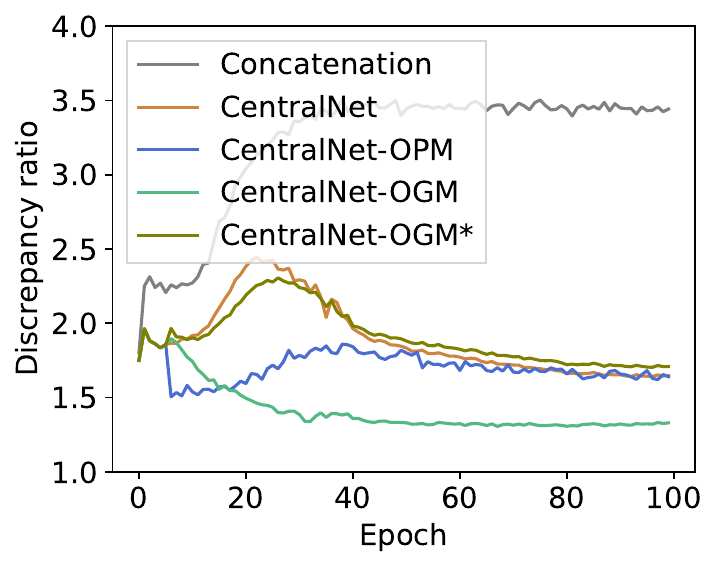}
    \label{fig:inter-ratio}
    }
    \subfigure[]{
    \includegraphics[width=0.46\linewidth]{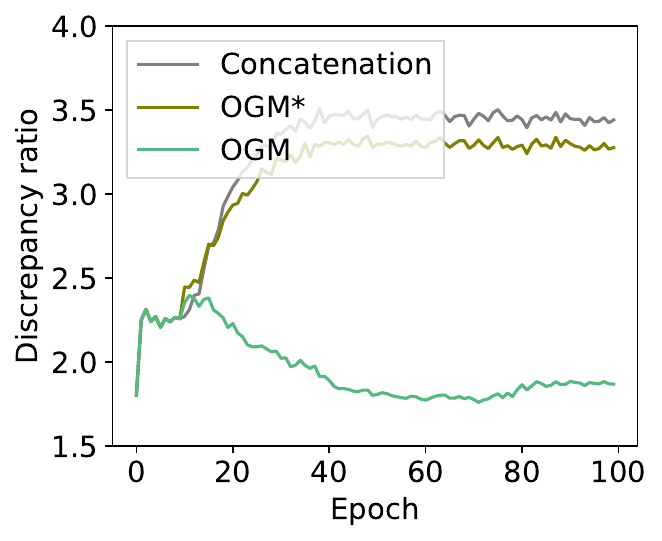}
    \label{fig:ratio-ogm}
    }

    \caption{\textbf{Discrepancy ratio on Kinetics-Sounds dataset.} (a) Discrepancy ratio of Concatenation, CentralNet\cite{vielzeuf2018centralnet}, and our methods during training. (b) Discrepancy ratio of Concatenation, OGM and OGM* methods based on Concatenation fusion during training. OGM* indicates the strategy that only increases the gradient of worse learnt modality. More discussion about OGM* is provided in~\autoref{sec:ogm_xing}.}
    \label{fig:ratio}
\end{figure}

\subsubsection{Imbalance modulation analysis} 
In~\autoref{fig-teaser}, we present the uni-modal and multimodal performance after applying our methods. It is observed that the performance of both audio and visual encoders in the multimodal models is enhanced following our modulation.  Additionally, an interesting observation is that our methods initially underperform compared to the baseline but ultimately surpass it. The reason could be that our modulation methods either drop the uni-modal feature or mitigate the gradient, decelerating the optimization of the modality with better performance at the beginning. And then the overall multimodal model exploits the information of another modality more and gains improvement in the end. To further analyze the modulation process, we monitor the change in discrepancy ratio between the two modalities during training on the Kinetics-Sounds dataset. The results are demonstrated in~\autoref{fig:ratio}. Based on the results, we can observe that the discrepancy ratio has an obvious drop after applying our modulation strategy, which provides concrete evidence for the effectiveness of our methods in alleviating the imbalanced learning problem. However, the two modalities are hard to have equal performance, \emph{i.e.,} the discrepancy ratio is hard to be close to 1, since the uni-modal discriminative information is naturally not equal in the curated dataset.

\subsubsection{Increase the gradient of worse learnt modality}
\label{sec:ogm_xing}
In OGM, we propose to mitigate the gradient of modality with better performance, and then the worse learnt modality can gain more training to ease the imbalanced learning problem. Corresponding to this idea of weakening the training of better modality, another idea is to accelerate the training of worse modality by increasing its gradient. To verify this opposite idea, we try the OGM* method, where the gradient of dominant modality remains unchanged, and the gradient of worse learnt modality is increased based on the uni-modal discrepancy. Experiments are conducted on CREMA-D and Kinetics-Sounds datasets. Based on the results in~\autoref{tab:ogm*}, we can find that compared with the Concatenation fusion method, OGM* method can also improve the model performance, but it still has an obvious performance gap compared with the original OGM. 

The reason could be that although OGM* accelerates the training of worse learnt modality by increasing its gradient, the training of more discriminative modality is not affected. More discriminative modality is still easy-to-learn for the multimodal model. Specifically, as shown in~\autoref{fig:ratio-ogm}, in OGM* method, the discriminative discrepancy ratio between modalities is still very large, which indicates the huge imbalance between modalities. The domination in multimodal optimization is not effectively weakened. Also, as shown in~\autoref{fig:inter-ratio}, OGM* method is hard to alleviate the performance discrepancy with the more complex CentralNet multimodal model. Thus, OGM* only has a limited performance improvement. However, in OGM method, the training of more discriminative modality is directly suppressed by mitigating its gradient. OGM directly makes the more discriminative modality harder to learn than before, accordingly weakening its domination. As~\autoref{fig:ratio-ogm}, with OGM, the discriminative discrepancy ratio between modalities is greatly decreased, which indicates the imbalance between modalities is successfully alleviated. In Appendix F, we provide more experiments about the OGM* strategy.

These results demonstrate that increasing the gradient of worse learnt modality could accelerate the optimization of the corresponding modality. However, the increased gradient may be not suitable at both direction and scale in the global view, leading to limited enhancement.

\begin{table}[t]
	\centering
 \caption{Experiments about OGM and OGM* strategies on CREMA-D and Kinetics-Sounds dataset. OGM* aims to increase the gradient of suppressed modality. All methods are based on Concatenation fusion.}
 \vspace{-1em}
	\begin{tabular}{c|cc|cc}
		\toprule
\multirow{2}{*}{\textbf{Method}} & \multicolumn{2}{c|}{\textbf{CREMA-D}} & \multicolumn{2}{c}{\textbf{Kinetics-Sounds}} \\
    & \textbf{Acc}      & \textbf{mAP}      & \textbf{Acc} & \multicolumn{1}{c}{\textbf{mAP}} \\
		\midrule[0.7pt]
		Concatenation & 66.9 & 72.1 & 63.0 & 67.3 \\
		OGM & \textbf{74.6} & \textbf{81.4} & \textbf{65.4} & \textbf{71.4} \\
            OGM* & 67.2 & 72.7 & 63.9 & 67.8 \\
        \bottomrule
	\end{tabular}
 \vspace{-1em}
\label{tab:ogm*}
\end{table}

\begin{figure*}[t]
    \centering
        \subfigure[Audio missing=$10\%$]{
        \includegraphics[width=0.23\linewidth]
        {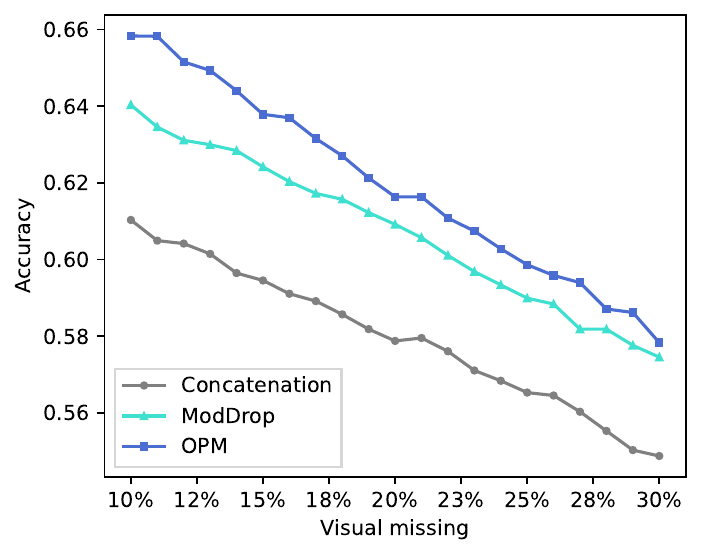}
        \label{fig:missing-audio}
        }
        \subfigure[Audio missing=$20\%$]{
        \includegraphics[width=0.23\linewidth]{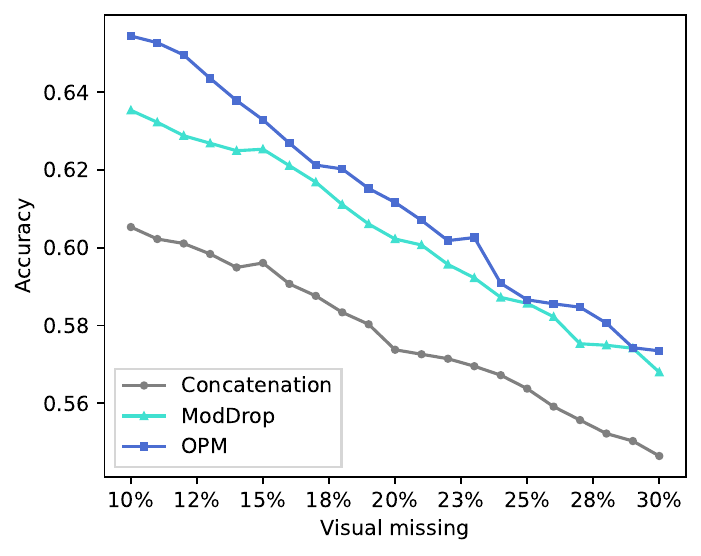}
        \label{fig:missing-audio02}
        }
        \subfigure[RGB missing=$10\%$]{
        \includegraphics[width=0.23\linewidth]{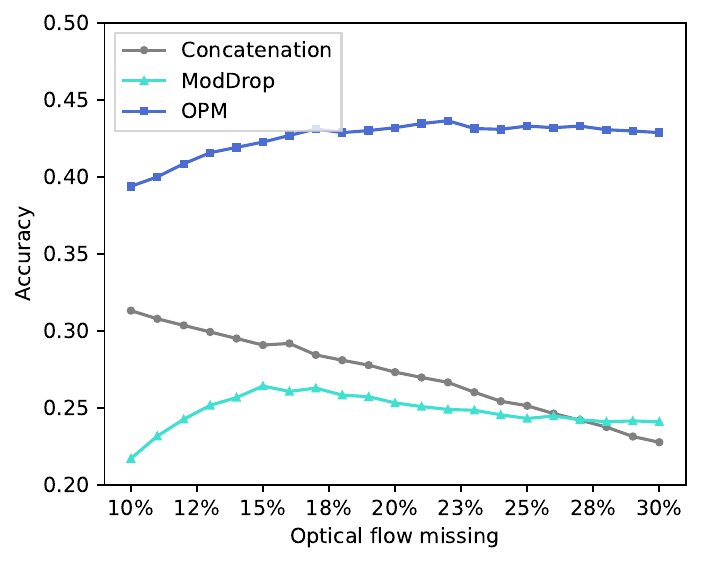}
        \label{fig7-1}
        }
        \subfigure[RGB missing=$20\%$]{
        \includegraphics[width=0.23\linewidth]{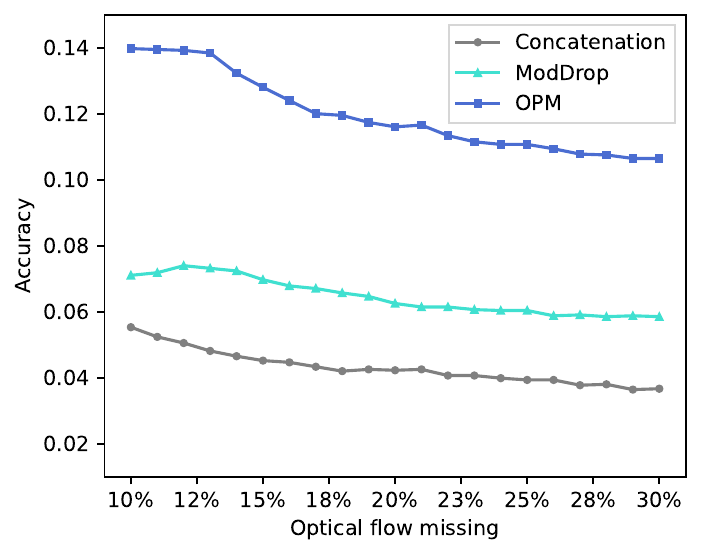}
        \label{fig7-2}
        }
    \caption{\textbf{(a\&b)}: Missing modality cases on Kinetics-Sounds dataset. \textbf{(c\&d)}: Missing modality cases on UCF-101 dataset. OPM method is based on Concatenation fusion.}
    \label{fig:missing}
    \vspace{-1em}
\end{figure*}

\subsubsection{Missing modality case of OPM}
It is not always available data on all modalities during testing, which is a quite common problem in the practice. Since the OPM method uses the modality dropout strategy, it is hopeful to defend this problem to some extent since the model has been trained on the missing modality case during the training process. Here we conduct missing modality evaluation for a trained multimodal model. During evaluation, for a trained multimodal model, each modality of testing samples is randomly dropped by specific probability. For example, in~\autoref{fig:missing-audio}, for testing samples of Kinetics-Sounds dataset, the audio modality is dropped with a probability $10\%$ and the visual modality is dropped with a probability ranging from $10\%$ to $30\%$. In practice, data of the dropped modality is set to $0$. The results are shown in~\autoref{fig:missing}. Besides the Concatenation baseline, the ModDrop~\cite{neverova2015moddrop} method that drops each modality with a fixed probability during training is also compared. Based on the results, the idea of modality dropout indeed can overcome the missing modality scenario to some extent. Both OPM and ModDrop are superior to the Concatenation baseline in most cases. Besides, our OPM with adaptive modality dropout probability further brings the improvement in model robustness.

In addition, one observation is that the performance of ModDrop and OPM methods does not keep dropping when the missing probability of optical flow modality is increased, in~\autoref{fig7-1}. This could be caused by the following reasons. For the UCF-101 dataset, RGB modality is clearly more discriminative than optical flow modality. As~\autoref{fig7-1} and~\autoref{fig7-2}, when the missing probability of RGB modality increases from $10\%$ to $20\%$, the accuracy of different methods drops dramatically. However, in contrast, as~\autoref{fig:missing-audio} and~\autoref{fig:missing-audio02}, when the missing probability of audio or visual modality increases, there is no such dramatic drop in accuracy. This demonstrates that the difference in discriminative ability between modalities on the UCF-101 dataset is more severe than others. For the UCF-101 dataset, multimodal models greatly depend on the more discriminative RGB modality even after modulation. But with the modulation of ModDrop and OPM methods, not only the ability of less discriminative optical flow modality, the ability of RGB modality in ModDrop and OPM is also further enhanced, compared with Concatenation baseline. Therefore, when the missing probability of RGB modality is low (10\% in~\autoref{fig7-1}), ModDrop and OPM methods with more powerful RGB encoders can still rely on the RGB modality to give correct predictions. Hence, their accuracy could even be marginally improved, when the noisy Optical-Flow modality is slightly absent. But as the missing percentage of Optical-Flow gradually increases, the accuracy of OPM and ModDrop begins to drop since too much Optical-Flow information is missing. When the missing probability of RGB modality is increased (20\% in~\autoref{fig7-2}), even for ModDrop and OPM methods, models can not purely rely on RGB modality, and the significance of optical flow modalities begins to stand out. The performance experiences a continuous drop when the missing probability of optical flow modality increases. We provide more experiments about the influence of noisy modality on missing modality cases in Appendix E.

\begin{figure}[t]
    \centering
    \subfigure[]{
    \includegraphics[width=0.46\linewidth]{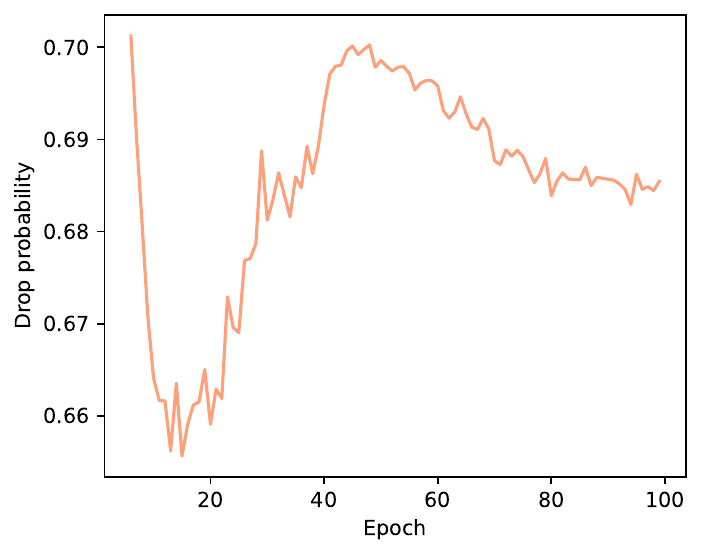}
    \label{fig:drop}
    }
    \subfigure[]{
    \includegraphics[width=0.46\linewidth]{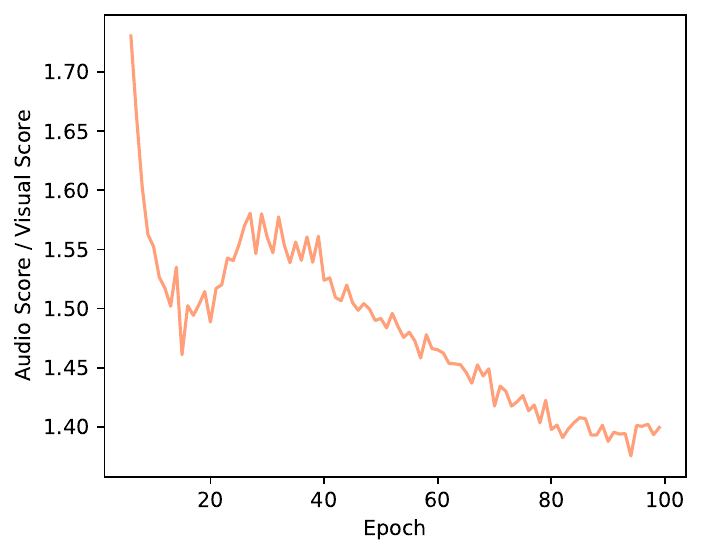}
    \label{fig:acc-dis}
    }
    \vspace{-1em} 
    \caption{(a) Drop probability of audio modality of OPM on Kinetics-Sounds dataset. $q_{base}=0.5$, $\lambda=0.5$. The drop probability of visual modality remains $0$. (b) Estimated discriminative score ratio of audio and visual modalities of OPM on Kinetics-Sounds dataset.}
    \vspace{-1em}
    \label{fig:opm-drop}
\end{figure}

\begin{figure*}[t]
    \centering
    \subfigure[playing saxophone]{
    \includegraphics[width=0.23\linewidth]{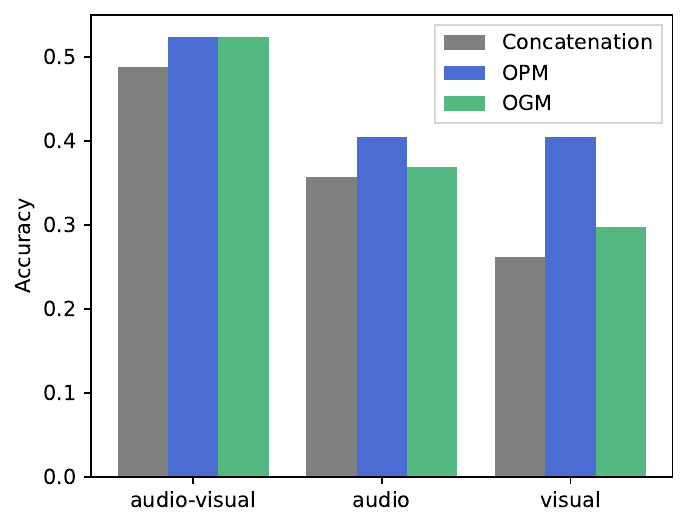}
    \label{category-wise-ks-04}
    }
    \subfigure[dribbling basketball]{
    \includegraphics[width=0.23\linewidth]{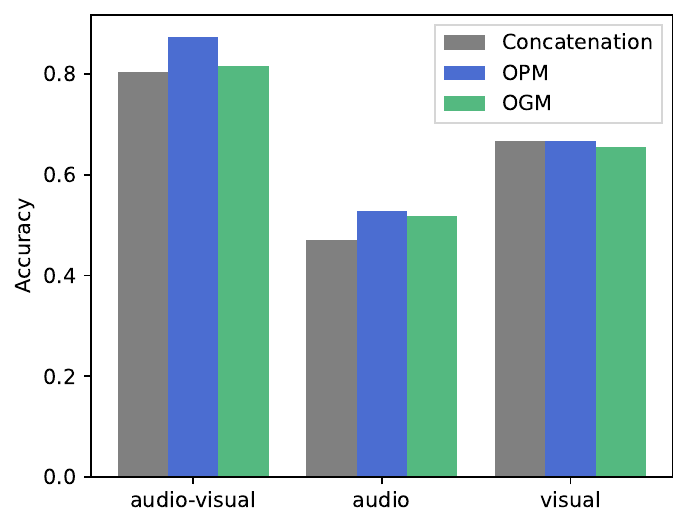}
    \label{category-wise-ks-05}
    }
    \subfigure[bathroom fan running]{
    \includegraphics[width=0.23\linewidth]{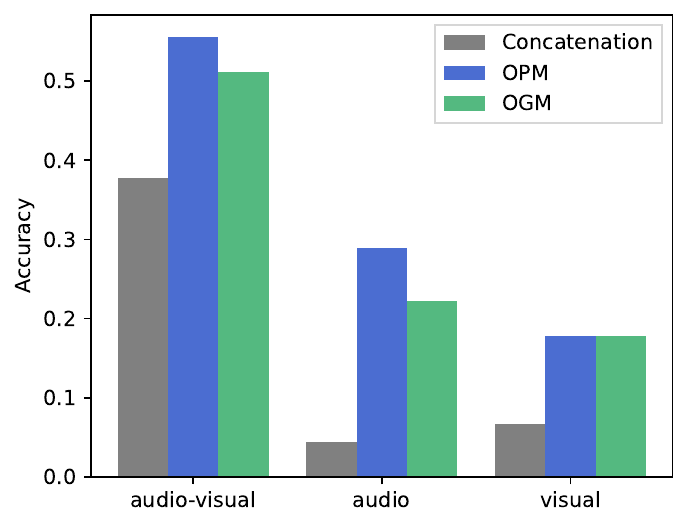}
    \label{category-wise-VGGSound-01}    
    }
    \subfigure[telephone bell ringing]{
    \includegraphics[width=0.23\linewidth]{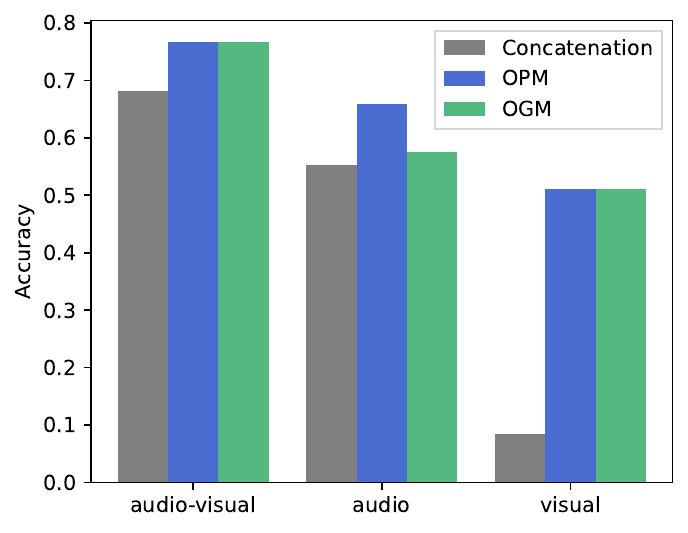}
    \label{category-wise-VGGSound-06}
    }
    \caption{\textbf{(a\&b)}: Category-wise analysis on Kinetics-Sounds dataset. \textbf{(c\&d)}: Category-wise analysis on VGGSound dataset. OPM and OGM methods are based on Concatenation fusion.}  
    \vspace{-1em}
    \label{fig:category-wise}
\end{figure*}

\subsubsection{Drop probability of each modality of OPM}
To further perceive the modulation of OPM during training, we record the specific changes in the modality dropout probability during training on the Kinetics-Sounds dataset. Results are shown in~\autoref{fig:drop}. In our OPM method, the drop probability of the modality with more discriminative information is adaptively adjusted during training, based on the degree of discriminative discrepancy between modalities. Hence, the ratio of estimated discriminative scores ($s$ in~\autoref{equ:uni-contribution}) between audio and visual modalities are also recorded and shown in~\autoref{fig:acc-dis}. Based on the results, the discriminative discrepancy between modalities changes dynamically during the training process, rather than remaining static. In addition, it can be seen that audio, as the dominant modality, always owns a higher accuracy than visual modality in the training process. Hence, the dropout probability for the audio modality is dynamic, while for the visual modality it remains at $0$ throughout training. The change in the estimated discriminative score ratio of audio and visual modalities reflects the need to adaptively adjust the dropout probabilities of each modality during training.

\subsubsection{Category-wise analysis}
To provide a fine-grained analysis of our methods, we conduct a category-wise analysis on the Kinetics-Sounds and VGGSound datasets. The results are shown in~\autoref{fig:category-wise}. Both our OPM and OGM methods improve the performance category-wise to a certain degree. Furthermore, we notice that the modality with less discriminative information generally achieves greater performance improvements when utilizing our methods. For example, the \textit{playing saxophone} category is more discriminative in audio and its audio modality earns less improvement. This phenomenon is consistent with our modulation which mitigates the influence of the dominant modality and then improves the learning of worse learnt modality. In addition, it is also validated that regardless of which modality dominates the training for a category, the OPM and OGM methods can effectively alleviate the imbalance. For instance, the \textit{dribbling basketball} and \textit{bathroom fan running} categories are more discriminative in visual (gray bar for audio is lower than visual in~\autoref{category-wise-ks-05} and ~\autoref{category-wise-VGGSound-01}), while the \textit{playing saxophone} and \textit{telephone bell ringing} categories have a preference in audio (gray bar for audio is higher than visual in~\autoref{category-wise-ks-04} and ~\autoref{category-wise-VGGSound-06}), and they both obtain enhancement.

Furthermore, although VGGSond is primarily a sound-oriented dataset with categories mainly determined by sound type, this does not mean that the visual information is non-discriminative. While the visual information is not directly related to the ``auditory label,'' it does reflect the corresponding sound events through the objects producing those sounds. The visual information of sounding objects can aid in distinguishing one sound type from another. Therefore, when our methods enhance the learning of the visual modality, the model’s classification ability improves due to the strengthened visual capability.

\begin{table}[]
\centering
\tabcolsep=3mm
\renewcommand\arraystretch{1.3}
\caption{Different split of bias term in the final classifier. All methods are based on Concatenation fusion.}
 \vspace{-1em}
\label{tab:bias}
\begin{tabular}{c|cc|cc}
\toprule
\textbf{Dataset} & \multicolumn{2}{c|}{\textbf{CREMA-D}} & \multicolumn{2}{c}{\textbf{Kinetics-Sounds}} \\ \midrule
Concatenation    & \multicolumn{2}{c|}{66.9}             & \multicolumn{2}{c}{63.0}                     \\ \midrule
Method           & OPM               & OGM               & OPM                   & OGM                  \\ \midrule
$\gamma^1=0$,$\gamma^2=0$        & 74.5              & 73.9              & 66.9                  & 65.4                 \\
$\gamma^1=1$,$\gamma^2=1$        & 74.2              & 74.7              & 66.7                  & 64.9                 \\ \midrule
$\gamma^1=\frac{1}{2}$,$\gamma^2=\frac{1}{2}$        & 75.1              & 74.6              & 67.0                  & 65.4                 \\
$\gamma^1=\frac{1}{3}$,$\gamma^2=\frac{2}{3}$        & 74.6              & 74.3              & 66.7                  & 65.1                 \\
$\gamma^1=\frac{2}{3}$,$\gamma^2=\frac{1}{3}$        & 73.7              & 74.5              & 66.5                  & 64.7                 \\ \bottomrule
\end{tabular}
\vspace{-1em}
\end{table}

\begin{table}[t]
\centering
\tabcolsep=0.5mm
 \caption{Accuracy of the variants of our modulation strategies on the Kinetics-Sounds and UCF-101 dataset. OPM and OGM methods are based on Concatenation fusion. KS denotes for Kinetics-Sounds.}
  \vspace{-1em}
\begin{tabular}{c|cc|cc}
\toprule
\textbf{Dataset}                                                                                              & \multicolumn{2}{c|}{\textbf{KS}}                       & \multicolumn{2}{c}{\textbf{UCF-101}}                    \\ \midrule
Concatenation                                                                                        & \multicolumn{2}{c|}{63.0}                     & \multicolumn{2}{c}{80.5}                      \\ \midrule
Method                                                                                               & OPM                   & OGM                   & OPM                   & OGM                   \\ \midrule
\multirow{2}{*}{$\rho^{m}_{t}= \frac{1}{M-1} \sum_{j\in[M],j\neq m} \frac{\sum_{i \in B_{t}}  s_{i}^{m} } {\sum_{i \in B_{t}}  s^{j}_i}$,} & \multirow{3}{*}{67.0} & \multirow{3}{*}{65.4} & \multirow{3}{*}{81.9} & \multirow{3}{*}{81.5} \\
                                                                                                     &                       &                       &                       &                       \\
$z(x)=tanh(x-1)$                                                                                     &                       &                       &                       &                       \\ \midrule
$\rho^{m}_{t}= \frac{1}{M-1} \sum_{j\in[M],j\neq m} \frac{\sum_{i \in B_{t}}  s_{i}^{m} } {\sum_{i \in B_{t}}  s^{j}_i}$,                  & \multirow{2}{*}{67.2} & \multirow{2}{*}{64.2} & \multirow{2}{*}{81.7} & \multirow{2}{*}{80.9} \\
$z(x)=sigmoid(x)$                                                                                    &                       &                       &                       &                       \\ \midrule
$\rho^{m}_{t}= \frac{1}{M-1} \sum_{j\in[M],j\neq m} (\sum_{i \in B_{t}}  s_{i}^{m} -  \sum_{i \in B_{t}}  s^{j}_i)$,                        & \multirow{2}{*}{66.9} & \multirow{2}{*}{65.4} & \multirow{2}{*}{81.8} & \multirow{2}{*}{80.8} \\
$z(x)=tanh(x-1)$                                                                                     &                       &                       &                       &                       \\ \midrule
$\rho^{m}_{t}= \frac{1}{M-1} \sum_{j\in[M],j\neq m} (\sum_{i \in B_{t}}  s_{i}^{m} -  \sum_{i \in B_{t}}  s^{j}_i)$,                        & \multirow{2}{*}{65.8} & \multirow{2}{*}{65.4} & \multirow{2}{*}{81.8} & \multirow{2}{*}{80.9} \\
$z(x)=sigmoid(x)$                                                                                    &                       &                       &                       &                       \\ \bottomrule
\end{tabular}
\vspace{-1em}
\label{tab:variants}
\end{table}

\subsection{Ablation study}

\subsubsection{Other variants of our modulation strategies}
\label{sec:variants}
Besides the strategy proposed in the method section, we also attempt several alternative strategies in this section. Notion follows~\autoref{sec:method}. Firstly, as stated in~\autoref{sec:method}, $s^{m}$ is the predicted probability for the correct category of modality $m$. Then, the total discriminative score of modality $m$ in a mini-batch is $\sum_{i \in B_{t}} s_{i}^{m}$. In~\autoref{sec:method}, the discriminative discrepancy degree among modalities is measured as the average ratio of this score, i.e., $\rho^{m}_{t}= \frac{1}{M-1} \sum_{j\in[M],j\neq m} \frac{\sum_{i \in B_{t}}  s_{i}^{m} } {\sum_{i \in B_{t}}  s^{j}_i}$. Here, we attempt to measure the discriminative discrepancy degree via the difference of this score, i.e., $\rho^{m}_{t}= \frac{1}{M-1} \sum_{j\in[M],j\neq m} (\sum_{i \in B_{t}}  s_{i}^{m} -  \sum_{i \in B_{t}}  s^{j}_i)$. 
Secondly, in the above experiments, $tanh(x-1)$ is used as the monotonically increasing function $z(x)$, which is used to map the discriminative discrepancy degree between modalities (\emph{i.e.,} $\rho_{t}$) into $(0,1)$ as the strength of modulation. We further attempt to use $sigmoid(x)$ function as $z(x)$ in this section. 
The results are shown in~\autoref{tab:variants}. We can have several observations. Firstly, the measurement of discriminative discrepancy degree between modalities does not greatly affect the effectiveness. Both the ratio of scores and the difference of scores can be used as the measure. Secondly, it does not need much effort on the specific selection of $z(x)$. Different combinations of $\rho_{t}$ and $z(x)$ can obtain consistency improvement compared with the Concatenation baseline across different datasets. Overall, these experiments indicate the effectiveness of the proposed modulation idea, demonstrating that our methods are not dependent on a specific design.

\subsubsection{Split of bias term in the final classifier}
\label{sec:bias}
In~\autoref{equ:uni-contribution}, when estimating the uni-modal discriminative performance for modality $m$, we use $W^{m}_{i} \cdot \varphi_{i}(\theta^{m},x_{i}^{m})+\frac{b}{M}$. The bias term in the final classifier is equally split. In this section, we try different splits of bias term: $W^{m}_{i} \cdot \varphi_{i}(\theta^{m},x_{i}^{m})+\gamma^m \cdot b$. Based on the results in~\autoref{tab:bias}, different splits of this bias term in the final classifier do not have a great influence on the effectiveness of our methods. The reason could be that the value of bias term is smaller than $W^{m}_{i} \cdot \varphi_{i}(\theta^{m},x_{i}^{m})$. Therefore, its split does not affect the uni-modal discriminative performance estimation a lot. Then, the effectiveness of our method has no reliance on the split of this bias term.

\subsubsection{Hyper-parameter sensitivity analysis}
\label{sec:hyper}
\begin{figure}[t]
    \centering
    \subfigure[$\lambda$ in OPM.]{
    \includegraphics[width=0.3\linewidth]{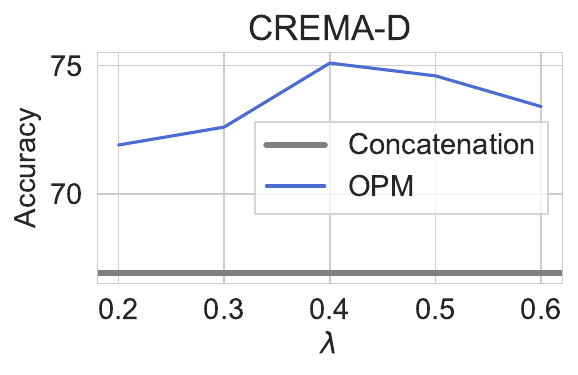}
    }
    \subfigure[$q_{base}$ in OPM.]{
    \includegraphics[width=0.3\linewidth]{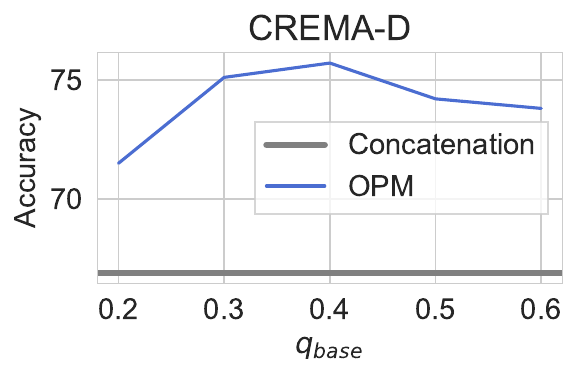}
    }
    \subfigure[$\alpha$ in OGM.]{
    \includegraphics[width=0.3\linewidth]{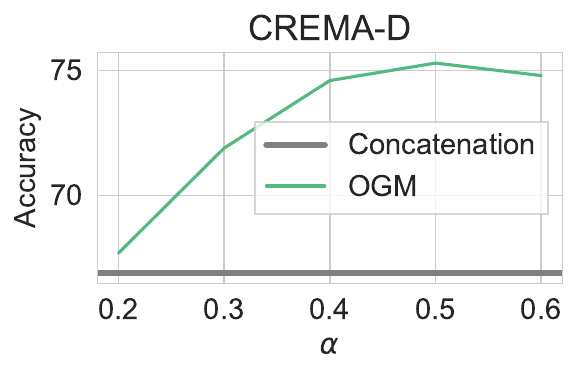}
    }

    \subfigure[$\lambda$ in OPM.]{
    \includegraphics[width=0.3\linewidth]{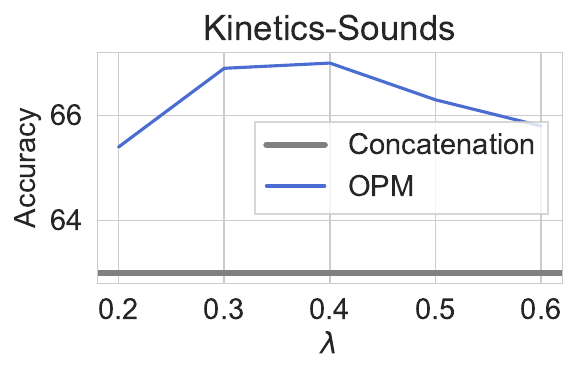}
    }
    \subfigure[$q_{base}$ in OPM.]{
    \includegraphics[width=0.3\linewidth]{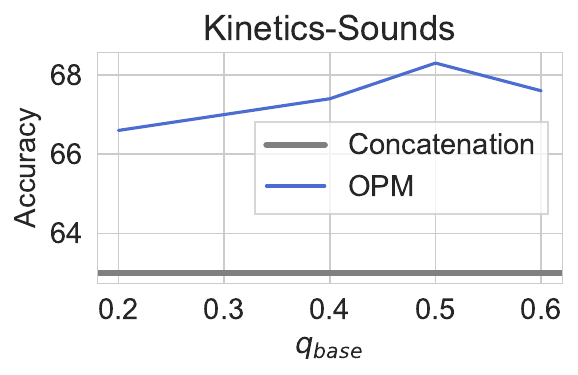}
    }
    \subfigure[$\alpha$ in OGM.]{
    \includegraphics[width=0.3\linewidth]{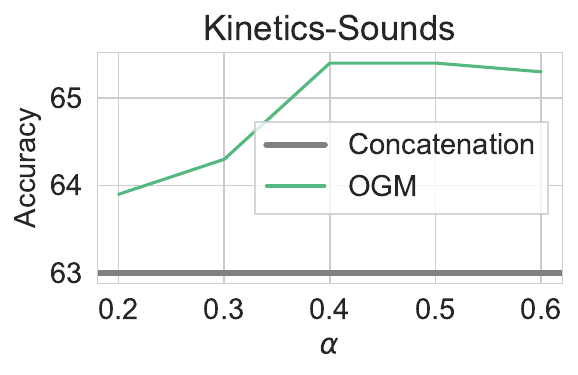}
    }
    \caption{Hyper-parameter sensitivity analysis of OPM and OGM on CREMA-D, Kinetics-Sounds dataset. OPM and OGM methods are based on Concatenation fusion.} 
    \vspace{-1em}
      \label{fig:sensitivity}
\end{figure}

In this section, we provide the hyper-parameter sensitivity analysis on the CREMA-D, Kinetics-Sounds and UCF-101 datasets. We select different values of $\lambda$ and $q_{base}$ in OPM as well as $\alpha$ in OGM. $q_{base}$ controls the lower bound and upper bound of modality dropout probability in OPM, while $\lambda$ and $\alpha$ control the degree of the modulation. The results are shown in~\autoref{fig:sensitivity}. According to the results, although the value of hyper-parameters with the best performance varies on different datasets, all selections of these three hyper-parameters can obtain consistent improvement compared with the Concatenation baseline. Therefore, it does not need much effort on the selection of hyper-parameters.

\section{Conclusion}
In this paper, we first observe and analyze the imbalanced learning phenomenon from both the feed-forward and the back-propagation stage in the multimodal learning, then propose on-the-fly modulation methods, OPM and OGM, to alleviate this problem. OPM mitigates the influence of the dominant modality by dropping its feature with dynamical probability in the feed-forward stage, while OGM weakens its gradient in the back-propagation stage. Both OPM and OGM aim to help the suppressed modality obtain more training. Moreover, our modulation methods are expected to have more sufficient learning of multimodal representation, since it enhances the learning of each modality. In the experiment, we combine the proposed methods with different multimodal models on various tasks and our methods achieve considerable improvement. A wide range of fine-grained analyses, several alternative variants and hyper-parameter sensitivity analyses are also provided to perceive their effectiveness from different views. Extensive experiment results demonstrate the promising effectiveness and versatility of the proposed methods.

\ifCLASSOPTIONcompsoc
  \section*{Acknowledgments}
\else
  \section*{Acknowledgment}
\fi

Zequn Yang participated in the theoretical
analysis discussion. We deeply thank him for his valuable assistance. This work was supported by National Natural Science Foundation of China (NO.62106272), the Young Elite Scientists Sponsorship Program by CAST (2021QNRC001), and Public Computing Cloud, Renmin University of China.

\ifCLASSOPTIONcaptionsoff
  \newpage
\fi



\bibliographystyle{IEEEtran}
\bibliography{refer.bib}

\begin{thebibliography}{10}
\providecommand{\url}[1]{#1}
\csname url@samestyle\endcsname
\providecommand{\newblock}{\relax}
\providecommand{\bibinfo}[2]{#2}
\providecommand{\BIBentrySTDinterwordspacing}{\spaceskip=0pt\relax}
\providecommand{\BIBentryALTinterwordstretchfactor}{4}
\providecommand{\BIBentryALTinterwordspacing}{\spaceskip=\fontdimen2\font plus
\BIBentryALTinterwordstretchfactor\fontdimen3\font minus
  \fontdimen4\font\relax}
\providecommand{\BIBforeignlanguage}[2]{{%
\expandafter\ifx\csname l@#1\endcsname\relax
\typeout{** WARNING: IEEEtran.bst: No hyphenation pattern has been}%
\typeout{** loaded for the language `#1'. Using the pattern for}%
\typeout{** the default language instead.}%
\else
\language=\csname l@#1\endcsname
\fi
#2}}
\providecommand{\BIBdecl}{\relax}
\BIBdecl

\bibitem{gazzaniga2006cognitive}
E.~B. Herreras, ``Cognitive neuroscience; the biology of the mind,''
  \emph{Cuadernos de Neuropsicolog{\'\i}a/Panamerican Journal of
  Neuropsychology}, vol.~4, no.~1, pp. 87--90, 2010.

\bibitem{wei2022learning}
Y.~Wei, D.~Hu, Y.~Tian, and X.~Li, ``Learning in audio-visual context: A
  review, analysis, and new perspective,'' \emph{arXiv preprint
  arXiv:2208.09579}, 2022.

\bibitem{kazakos2019epic}
E.~Kazakos, A.~Nagrani, A.~Zisserman, and D.~Damen, ``Epic-fusion: Audio-visual
  temporal binding for egocentric action recognition,'' in \emph{ICCV}, 2019.

\bibitem{gao2020listen}
R.~Gao, T.-H. Oh, K.~Grauman, and L.~Torresani, ``Listen to look: Action
  recognition by previewing audio,'' in \emph{CVPR}, 2020.

\bibitem{choudhury2018segmentation}
A.~R. Choudhury, R.~Vanguri, S.~R. Jambawalikar, and P.~Kumar, ``Segmentation
  of brain tumors using deeplabv3+,'' in \emph{International MICCAI Brainlesion
  Workshop}, 2018.

\bibitem{cao2021shapeconv}
J.~Cao, H.~Leng, D.~Lischinski, D.~Cohen-Or, C.~Tu, and Y.~Li, ``Shapeconv:
  Shape-aware convolutional layer for indoor rgb-d semantic segmentation,''
  \emph{arXiv preprint arXiv:2108.10528}, 2021.

\bibitem{tian2018audio}
Y.~Tian, J.~Shi, B.~Li, Z.~Duan, and C.~Xu, ``Audio-visual event localization
  in unconstrained videos,'' in \emph{ECCV}, 2018.

\bibitem{wang2020makes}
W.~Wang, D.~Tran, and M.~Feiszli, ``What makes training multi-modal
  classification networks hard?'' in \emph{CVPR}, 2020.

\bibitem{sun2021learning}
Y.~Sun, S.~Mai, and H.~Hu, ``Learning to balance the learning rates between
  various modalities via adaptive tracking factor,'' \emph{IEEE Signal
  Processing Letters}, vol.~28, pp. 1650--1654, 2021.

\bibitem{huang2022modality}
Y.~Huang, J.~Lin, C.~Zhou, H.~Yang, and L.~Huang, ``Modality competition: What
  makes joint training of multi-modal network fail in deep
  learning?(provably),'' \emph{arXiv preprint arXiv:2203.12221}, 2022.

\bibitem{wu2022characterizing}
N.~Wu, S.~Jastrzebski, K.~Cho, and K.~J. Geras, ``Characterizing and overcoming
  the greedy nature of learning in multi-modal deep neural networks,'' in
  \emph{ICML}, 2022.

\bibitem{vielzeuf2018centralnet}
V.~Vielzeuf, A.~Lechervy, S.~Pateux, and F.~Jurie, ``Centralnet: a multilayer
  approach for multimodal fusion,'' in \emph{ECCV Workshops}, 2018.

\bibitem{chen2020vggsound}
H.~Chen, W.~Xie, A.~Vedaldi, and A.~Zisserman, ``Vggsound: A large-scale
  audio-visual dataset,'' in \emph{ICASSP}, 2020.

\bibitem{parida2020coordinated}
K.~Parida, N.~Matiyali, T.~Guha, and G.~Sharma, ``Coordinated joint multimodal
  embeddings for generalized audio-visual zero-shot classification and
  retrieval of videos,'' in \emph{WACV}, 2020.

\bibitem{peng2022balanced}
X.~Peng, Y.~Wei, A.~Deng, D.~Wang, and D.~Hu, ``Balanced multimodal learning
  via on-the-fly gradient modulation,'' in \emph{CVPR}, 2022.

\bibitem{simonyan2014two}
K.~Simonyan and A.~Zisserman, ``Two-stream convolutional networks for action
  recognition in videos,'' \emph{NeurIPS}, 2014.

\bibitem{potamianos2004audio}
G.~Potamianos, C.~Neti, J.~Luettin, and I.~Matthews, ``Audio-visual automatic
  speech recognition: An overview,'' \emph{Issues in visual and audio-visual
  speech processing}, vol.~22, p.~23, 2004.

\bibitem{hu2016temporal}
D.~Hu, X.~Li \emph{et~al.}, ``Temporal multimodal learning in audiovisual
  speech recognition,'' in \emph{CVPR}, 2016.

\bibitem{lin2019dual}
Y.-B. Lin, Y.-J. Li, and Y.-C.~F. Wang, ``Dual-modality seq2seq network for
  audio-visual event localization,'' in \emph{ICASSP}, 2019.

\bibitem{ilievski2017multimodal}
I.~Ilievski and J.~Feng, ``Multimodal learning and reasoning for visual
  question answering,'' in \emph{NeurIPS}, 2017.

\bibitem{winterbottom2020modality}
T.~Winterbottom, S.~Xiao, A.~McLean, and N.~A. Moubayed, ``On modality bias in
  the tvqa dataset,'' \emph{arXiv preprint arXiv:2012.10210}, 2020.

\bibitem{du2021improving}
C.~Du, T.~Li, Y.~Liu, Z.~Wen, T.~Hua, Y.~Wang, and H.~Zhao, ``Improving
  multi-modal learning with uni-modal teachers,'' \emph{arXiv preprint
  arXiv:2106.11059}, 2021.

\bibitem{wei2024innocent}
Y.~Wei and D.~Hu, ``Mmpareto: boosting multimodal learning with innocent
  unimodal assistance,'' in \emph{ICML}, 2024.

\bibitem{wei2024enhancing}
Y.~Wei, R.~Feng, Z.~Wang, and D.~Hu, ``Enhancing multimodal cooperation via
  sample-level modality valuation,'' in \emph{CVPR}, 2024.

\bibitem{wei2024diagnosing}
Y.~Wei, S.~Li, R.~Feng, and D.~Hu, ``Diagnosing and re-learning for balanced
  multimodal learning,'' in \emph{ECCV}, 2024.

\bibitem{yang2024Quantifying}
Z.~Yang, Y.~Wei, C.~Liang, and D.~Hu, ``Quantifying and enhancing multi-modal
  robustness with modality preference,'' \emph{arXiv preprint
  arXiv:2402.06244}, 2024.

\bibitem{srivastava2014dropout}
N.~Srivastava, G.~Hinton, A.~Krizhevsky, I.~Sutskever, and R.~Salakhutdinov,
  ``Dropout: a simple way to prevent neural networks from overfitting,''
  \emph{The journal of machine learning research}, vol.~15, no.~1, pp.
  1929--1958, 2014.

\bibitem{rennie2014annealed}
S.~J. Rennie, V.~Goel, and S.~Thomas, ``Annealed dropout training of deep
  networks,'' in \emph{SLT Workshop}, 2014.

\bibitem{li2016improved}
Z.~Li, B.~Gong, and T.~Yang, ``Improved dropout for shallow and deep
  learning,'' \emph{NeurIPS}, 2016.

\bibitem{morerio2017curriculum}
P.~Morerio, J.~Cavazza, R.~Volpi, R.~Vidal, and V.~Murino, ``Curriculum
  dropout,'' in \emph{ICCV}, 2017.

\bibitem{neverova2015moddrop}
N.~Neverova, C.~Wolf, G.~Taylor, and F.~Nebout, ``Moddrop: adaptive multi-modal
  gesture recognition,'' \emph{IEEE Transactions on Pattern Analysis and
  Machine Intelligence}, vol.~38, no.~8, pp. 1692--1706, 2015.

\bibitem{li2016multi}
X.~Li, Q.~Dou, H.~Chen, C.-W. Fu, and P.-A. Heng, ``Multi-scale and modality
  dropout learning for intervertebral disc localization and segmentation,'' in
  \emph{International Workshop on Computational Methods and Clinical
  Applications for Spine Imaging}, 2016, pp. 85--91.

\bibitem{hussen2020modality}
A.~Hussen~Abdelaziz, B.-J. Theobald, P.~Dixon, R.~Knothe, N.~Apostoloff, and
  S.~Kajareker, ``Modality dropout for improved performance-driven talking
  faces,'' in \emph{ICMI}, 2020.

\bibitem{de2020input}
S.~de~Blois, M.~Garon, C.~Gagn{\'e}, and J.-F. Lalonde, ``Input dropout for
  spatially aligned modalities,'' in \emph{ICIP}, 2020.

\bibitem{xiao2020audiovisual}
F.~Xiao, Y.~J. Lee, K.~Grauman, J.~Malik, and C.~Feichtenhofer, ``Audiovisual
  slowfast networks for video recognition,'' \emph{arXiv preprint
  arXiv:2001.08740}, 2020.

\bibitem{zhu2019anisotropic}
Z.~Zhu, J.~Wu, B.~Yu, L.~Wu, and J.~Ma, ``The anisotropic noise in stochastic
  gradient descent: Its behavior of escaping from sharp minima and
  regularization effects,'' in \emph{ICML}, 2019.

\bibitem{chaudhari2018stochastic}
P.~Chaudhari and S.~Soatto, ``Stochastic gradient descent performs variational
  inference, converges to limit cycles for deep networks,'' in \emph{ITA
  Workshop}, 2018.

\bibitem{xie2021artificial}
Z.~Xie, F.~He, S.~Fu, I.~Sato, D.~Tao, and M.~Sugiyama, ``Artificial neural
  variability for deep learning: On overfitting, noise memorization, and
  catastrophic forgetting,'' \emph{Neural computation}, vol.~33, no.~8, pp.
  2163--2192, 2021.

\bibitem{wu2020noisy}
J.~Wu, W.~Hu, H.~Xiong, J.~Huan, V.~Braverman, and Z.~Zhu, ``On the noisy
  gradient descent that generalizes as sgd,'' in \emph{ICML}, 2020.

\bibitem{he2019control}
F.~He, T.~Liu, and D.~Tao, ``Control batch size and learning rate to generalize
  well: Theoretical and empirical evidence,'' \emph{NeurIPS}, 2019.

\bibitem{jastrzkebski2017three}
S.~Jastrz{\k{e}}bski, Z.~Kenton, D.~Arpit, N.~Ballas, A.~Fischer, Y.~Bengio,
  and A.~Storkey, ``Three factors influencing minima in sgd,'' \emph{arXiv
  preprint arXiv:1711.04623}, 2017.

\bibitem{jin2017escape}
C.~Jin, R.~Ge, P.~Netrapalli, S.~M. Kakade, and M.~I. Jordan, ``How to escape
  saddle points efficiently,'' in \emph{ICML}, 2017.

\bibitem{neelakantan2015adding}
A.~Neelakantan, L.~Vilnis, Q.~V. Le, I.~Sutskever, L.~Kaiser, K.~Kurach, and
  J.~Martens, ``Adding gradient noise improves learning for very deep
  networks,'' \emph{stat}, vol. 1050, p.~21, 2015.

\bibitem{zhou2019toward}
M.~Zhou, T.~Liu, Y.~Li, D.~Lin, E.~Zhou, and T.~Zhao, ``Toward understanding
  the importance of noise in training neural networks,'' in \emph{ICML}, 2019.

\bibitem{wei2020implicit}
C.~Wei, S.~Kakade, and T.~Ma, ``The implicit and explicit regularization
  effects of dropout,'' in \emph{ICML}, 2020.

\bibitem{perez2018film}
E.~Perez, F.~Strub, H.~De~Vries, V.~Dumoulin, and A.~Courville, ``Film: Visual
  reasoning with a general conditioning layer,'' in \emph{AAAI}, 2018.

\bibitem{cao2014crema}
H.~Cao, D.~G. Cooper, M.~K. Keutmann, R.~C. Gur, A.~Nenkova, and R.~Verma,
  ``Crema-d: Crowd-sourced emotional multimodal actors dataset,'' \emph{IEEE
  transactions on affective computing}, vol.~5, no.~4, pp. 377--390, 2014.

\bibitem{arandjelovic2017look}
R.~Arandjelovic and A.~Zisserman, ``Look, listen and learn,'' in \emph{CVPR},
  2017.

\bibitem{kay2017kinetics}
W.~Kay, J.~Carreira, K.~Simonyan, B.~Zhang, C.~Hillier, S.~Vijayanarasimhan,
  F.~Viola, T.~Green, T.~Back, P.~Natsev \emph{et~al.}, ``The kinetics human
  action video dataset,'' \emph{arXiv preprint arXiv:1705.06950}, 2017.

\bibitem{soomro2012UCF101}
K.~Soomro, A.~R. Zamir, and M.~Shah, ``Ucf101: A dataset of 101 human actions
  classes from videos in the wild,'' \emph{arXiv preprint arXiv:1212.0402},
  2012.

\bibitem{zadeh2016mosi}
A.~Zadeh, R.~Zellers, E.~Pincus, and L.-P. Morency, ``Mosi: multimodal corpus
  of sentiment intensity and subjectivity analysis in online opinion videos,''
  \emph{arXiv preprint arXiv:1606.06259}, 2016.

\bibitem{li2022learning}
G.~Li, Y.~Wei, Y.~Tian, C.~Xu, J.-R. Wen, and D.~Hu, ``Learning to answer
  questions in dynamic audio-visual scenarios,'' in \emph{CVPR}, 2022.

\bibitem{he2016deep}
K.~He, X.~Zhang, S.~Ren, and J.~Sun, ``Deep residual learning for image
  recognition,'' in \emph{CVPR}, 2016.

\bibitem{zhao2018sound}
H.~Zhao, C.~Gan, A.~Rouditchenko, C.~Vondrick, J.~McDermott, and A.~Torralba,
  ``The sound of pixels,'' in \emph{ECCV}, 2018.

\bibitem{akbari2021vatt}
H.~Akbari, L.~Yuan, R.~Qian, W.-H. Chuang, S.-F. Chang, Y.~Cui, and B.~Gong,
  ``Vatt: Transformers for multimodal self-supervised learning from raw video,
  audio and text,'' \emph{NeurIPS}, 2021.

\bibitem{joze2020mmtm}
H.~R.~V. Joze, A.~Shaban, M.~L. Iuzzolino, and K.~Koishida, ``Mmtm: Multimodal
  transfer module for cnn fusion,'' in \emph{CVPR}, 2020.

\bibitem{zhou2021positive}
J.~Zhou, L.~Zheng, Y.~Zhong, S.~Hao, and M.~Wang, ``Positive sample propagation
  along the audio-visual event line,'' in \emph{CVPR}, 2021.

\bibitem{schwartz2019simple}
I.~Schwartz, A.~G. Schwing, and T.~Hazan, ``A simple baseline for audio-visual
  scene-aware dialog,'' in \emph{CVPR}, 2019.

\bibitem{yun2021pano}
H.~Yun, Y.~Yu, W.~Yang, K.~Lee, and G.~Kim, ``Pano-avqa: Grounded audio-visual
  question answering on 360deg videos,'' in \emph{ICCV}, 2021.

\bibitem{heracleous2019comprehensive}
P.~Heracleous and A.~Yoneyama, ``A comprehensive study on bilingual and
  multilingual speech emotion recognition using a two-pass classification
  scheme,'' \emph{PloS one}, vol.~14, no.~8, p. e0220386, 2019.

\bibitem{pappagari2020x}
R.~Pappagari, T.~Wang, J.~Villalba, N.~Chen, and N.~Dehak, ``x-vectors meet
  emotions: A study on dependencies between emotion and speaker recognition,''
  in \emph{ICASSP}, 2020.

\bibitem{ghaleb2019multimodal}
E.~Ghaleb, M.~Popa, and S.~Asteriadis, ``Multimodal and temporal perception of
  audio-visual cues for emotion recognition,'' in \emph{ACII}, 2019.

\bibitem{wang2016temporal}
L.~Wang, Y.~Xiong, Z.~Wang, Y.~Qiao, D.~Lin, X.~Tang, and L.~Van~Gool,
  ``Temporal segment networks: Towards good practices for deep action
  recognition,'' in \emph{ECCV}, 2016.

\bibitem{lin2019tsm}
J.~Lin, C.~Gan, and S.~Han, ``Tsm: Temporal shift module for efficient video
  understanding,'' in \emph{ICCV}, 2019.

\bibitem{ghorbani2020neuron}
A.~Ghorbani and J.~Y. Zou, ``Neuron shapley: Discovering the responsible
  neurons,'' \emph{NeurIPS}, 2020.

\bibitem{hu2022shape}
P.~Hu, X.~Li, and Y.~Zhou, ``Shape: An unified approach to evaluate the
  contribution and cooperation of individual modalities,'' \emph{IJCAI}, 2022.

\end{thebibliography}
%

%




\end{document}